\documentclass{article}
\usepackage{authblk}
\usepackage[utf8]{inputenc}
\usepackage{main}
\usepackage{microtype}
\usepackage{subcaption}
\usepackage{graphicx}
\usepackage{times}
\usepackage{latexsym}
\usepackage{amsmath}
\usepackage{float}
\usepackage{booktabs}
\usepackage{footnote}
\usepackage{enumitem}
\usepackage{bm}
\usepackage{arydshln}
\usepackage{booktabs}
\usepackage{multicol}
\usepackage{multirow}
\usepackage{color}
\usepackage[dvipsnames]{xcolor}     
\usepackage{colortbl}
\usepackage{bbding}
\usepackage{makecell}
\usepackage{mathtools}
\usepackage{imakeidx}
\usepackage{longtable}
\usepackage{wrapfig}
\usepackage{rotating}
\makeindex
\usepackage{arydshln}
\usepackage{lipsum}
\usepackage{natbib}
\usepackage[toc]{multitoc}
\usepackage[edges]{forest}
\usepackage[normalem]{ulem}
\definecolor{mydarkblue}{rgb}{0,0.08,0.45}
\usepackage[colorlinks=true,linkcolor=mydarkblue,citecolor=mydarkblue,filecolor=mydarkblue,urlcolor=mydarkblue]{hyperref}
\usepackage{CJKutf8}
\usepackage{awesomebox} 
\usepackage{bbding}
\usepackage[most]{tcolorbox}
\usepackage{geometry}
\definecolor{wkblue}{rgb}{0.2, 0.3, 0.6}
\definecolor{meta-color}{rgb}{0.5, 0.5, 0.5}
\usepackage{amsmath}
\usepackage{enumitem}
\usepackage{lscape} 
\usepackage{algorithm}    

\usepackage{algorithmicx}
\usepackage{algpseudocode}
\usepackage{tabularx}
\usepackage{makecell}

\usepackage{amssymb}
\usepackage{amsfonts}

\usepackage[tikz]{bclogo}
\usepackage[framemethod=tikz]{mdframed}
\definecolor{bgblue}{RGB}{245,243,253}
\definecolor{ttblue}{RGB}{91,194,224}

\mdfdefinestyle{mystyle}{%
  rightline=true,
  innerleftmargin=10,
  innerrightmargin=10,
  outerlinewidth=3pt,
  topline=false,
  rightline=true,
  bottomline=false,
  skipabove=\topsep,
  skipbelow=\topsep
}

\usepackage[utf8]{inputenc}
\usepackage{amsmath, amssymb}
\usepackage{tcolorbox}
\usepackage{listings}
\usepackage{geometry}
\definecolor{codegray}{RGB}{248,248,248}
\definecolor{myblue}{RGB}{0,32,96}
\definecolor{darkgreen}{RGB}{0,100,0}
\lstdefinestyle{mypython}{
    language=Python,
    backgroundcolor=\color{white},
    basicstyle=\ttfamily\scriptsize,
    frame=single,
    breaklines=true,
    keywordstyle=\color{blue},
    commentstyle=\color{gray},
    showstringspaces=false,
}

\newtcolorbox{myboxi}[1][]{
  breakable,
  title=#1,
  colback=red!5,
  colbacktitle=red!5,
  coltitle=black,
  fonttitle=\bfseries,
  bottomrule=0pt,
  toprule=0pt,
  leftrule=2pt,
  rightrule=2pt,
  titlerule=0pt,
  arc=0pt,
  outer arc=0pt,
  colframe=red,
}

\newtcolorbox{myboxnote}[1][]{
  breakable,
  title=#1,
  colback=orange!0,
  colbacktitle=orange!0,
  coltitle=black,
  fonttitle=\bfseries,
  bottomrule=0pt,
  toprule=0pt,
  leftrule=2pt,
  rightrule=2pt,
  titlerule=0pt,
  arc=0pt,
  outer arc=0pt,
  colframe=orange,
}

\newtcolorbox{myboxii}[1][]{
  breakable,
  freelance,
  title=#1,
  colback=white,
  colbacktitle=white,
  coltitle=black,
  fonttitle=\bfseries,
  bottomrule=0pt,
  boxrule=0pt,
  colframe=white,
  overlay unbroken and first={
  \draw[red!75!black,line width=3pt]
    ([xshift=5pt]frame.north west) -- 
    (frame.north west) -- 
    (frame.south west);
  \draw[red!75!black,line width=3pt]
    ([xshift=-5pt]frame.north east) -- 
    (frame.north east) -- 
    (frame.south east);
  },
  overlay unbroken app={
  \draw[red!75!black,line width=3pt,line cap=rect]
    (frame.south west) -- 
    ([xshift=5pt]frame.south west);
  \draw[red!75!black,line width=3pt,line cap=rect]
    (frame.south east) -- 
    ([xshift=-5pt]frame.south east);
  },
  overlay middle and last={
  \draw[red!75!black,line width=3pt]
    (frame.north west) -- 
    (frame.south west);
  \draw[red!75!black,line width=3pt]
    (frame.north east) -- 
    (frame.south east);
  },
  overlay last app={
  \draw[red!75!black,line width=3pt,line cap=rect]
    (frame.south west) --
    ([xshift=5pt]frame.south west);
  \draw[red!75!black,line width=3pt,line cap=rect]
    (frame.south east) --
    ([xshift=-5pt]frame.south east);
  },
}

\usepackage{fancyhdr} 
\usepackage{blindtext} 
\usepackage{makecell}

\DeclareCaptionFont{black}{\color{black}}

\newcommand{\Thead}[1]{\textbf{\textsc{#1}}}
\newcommand{\highlight}[1]{\textbf{\textcolor{red}{#1}}}

\newcommand{\examplequestion}[1]{\textbf{\textcolor{darkgreen}{#1}}}

\definecolor{myblue}{rgb}{0.9, 0.1, 0.94}
\definecolor{mygreen}{rgb}{0.64, 0.56, 0.88}
\definecolor{myyellow}{rgb}{0.68, 0.6, 0.1}
\definecolor{fancygreen}{rgb}{0.33, 0.68, 0.20}
\definecolor{salmon}{rgb}{0.94, 0.52, 0.49}
\definecolor{tablegreen}{rgb}{0.82, 0.94, 0.75}
\definecolor{tableblue}{rgb}{0.81, 0.90, 0.94}
\definecolor{tablered}{rgb}{0.97, 0.85, 0.85}
\definecolor{tableorange}{rgb}{0.96, 0.85, 0.81}
\definecolor{lightblue}{RGB}{173, 216, 230}
\definecolor{SampleSeparator}{rgb}{0.96, 0.85, 0.81}

\newenvironment{itemize*}%
 {\leftmargini=10pt\begin{itemize}%
  \setlength{\itemsep}{0pt}%
  \setlength{\parskip}{0pt}%
  }%
 {\end{itemize}}
\newenvironment{enumerate*}%
 {\begin{enumerate}%
  \setlength{\itemsep}{0pt}%
  \setlength{\parskip}{0pt}}%
 {\end{enumerate}}

\usepackage{listings}

\newcommand\JSONnumbervaluestyle{\color{blue}}
\newcommand\JSONstringvaluestyle{\color{red}}

\newif\ifcolonfoundonthisline

\makeatletter

\lstdefinestyle{json}
{
  showstringspaces    = false,
  keywords            = {false,true},
  alsoletter          = 0123456789.,
  morestring          = [s]{"}{"},
  stringstyle         = \ifcolonfoundonthisline\JSONstringvaluestyle\fi,
  MoreSelectCharTable =%
    \lst@DefSaveDef{`:}\colon@json{\processColon@json},
  basicstyle          = \ttfamily,
  keywordstyle        = \ttfamily\bfseries,
}

\newcommand\processColon@json{%
  \colon@json%
  \ifnum\lst@mode=\lst@Pmode%
    \global\colonfoundonthislinetrue%
  \fi
}

\lst@AddToHook{Output}{%
  \ifcolonfoundonthisline%
    \ifnum\lst@mode=\lst@Pmode%
      \def\lst@thestyle{\JSONnumbervaluestyle}%
    \fi
  \fi
  \lsthk@DetectKeywords%
}

\lst@AddToHook{EOL}%
  {\global\colonfoundonthislinefalse}

\makeatother

\usepackage{etoolbox}
\usepackage{natbib}
\usepackage{url}
\newcounter{bibcount}
\makeatletter
\patchcmd{\@lbibitem}{\item[}{\item[\hfil\stepcounter{bibcount}{[\thebibcount]}}{}{}
\renewcommand\NAT@bibsetup%
  [1]{\setlength{\leftmargin}{\bibhang}\setlength{\itemindent}{-\parindent}%
      \setlength{\itemsep}{\bibsep}\setlength{\parsep}{\z@}}
\makeatother

\definecolor{mybrown}{RGB}{128,64,0}
\definecolor{titlecolor}{HTML}{4c9cff}
\newcommand{\gair}{$^{\color{Emerald}{\gamma}}$}
\newcommand{\sjtu}{$^{\color{RoyalBlue}{\sigma}}$}
\newcommand{\alibaba}{$^{\color{BurntOrange}{\tau}}$}

\newcommand{\syntheticprime}{\textit{Synthetic Prime }}
\newcommand{\syntheticprimenospace}{\textit{Synthetic Prime}}

\fancypagestyle{normal}{
  \cfoot{\thepage}
  \lhead{\includegraphics[height=0.67cm]{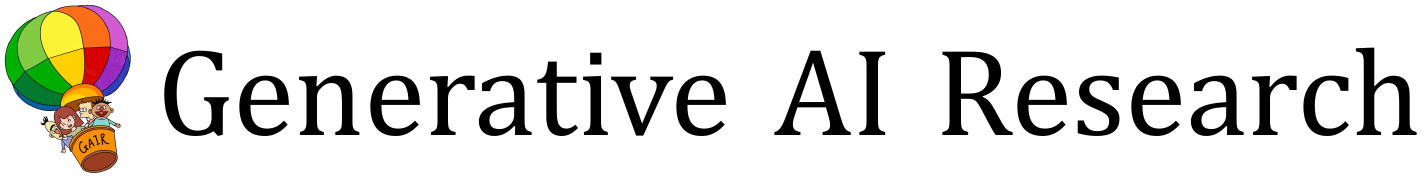}}
  
  \setlength{\headheight}{15pt}
  \setlength{\headsep}{5mm}
}

\fancypagestyle{titlepage}{
  \cfoot{\thepage}
  \lhead{\includegraphics[height=0.67cm]{assets/GAIR_Logo3.pdf}}
  
  \setlength{\headheight}{15pt}
  \setlength{\headsep}{0mm}
}

\makeatletter
\let\ps@plain\ps@normal
\makeatother

\begin{document}

\title{\includegraphics[width=0.06\textwidth]{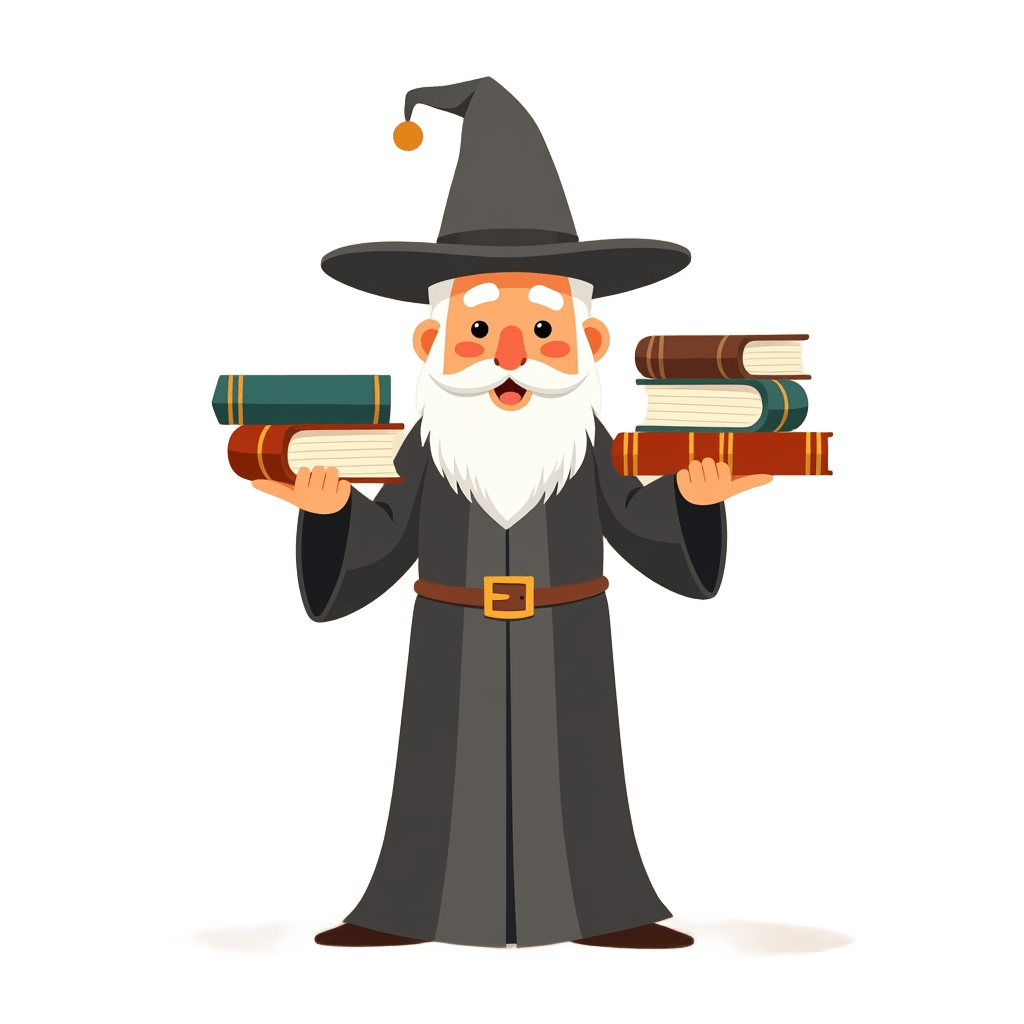}One Sample to Rule Them All: Extreme Data Efficiency in Multidiscipline Reasoning with Reinforcement Learning}

\author{
\textbf{Yiyuan Li\gair\space\space\space Zhen Huang\gair\space\space\space Yanan Wu\alibaba \space\space\space Weixun Wang\alibaba} \\
\vspace{-1em}
\textbf{Xuefeng Li\gair\space\space\space 
Yijia Luo\alibaba\space\space\space 
Wenbo Su\alibaba\space\space\space Bo Zheng\alibaba\space\space\space Pengfei Liu\sjtu\gair\thanks{Corresponding author}}

Taobao \& Tmall Group of Alibaba \alibaba, Shanghai Jiaotong Univeristy\sjtu, GAIR\gair\\
\texttt{yiyuanli@cs.unc.edu}\\
\includegraphics[width=0.02\textwidth]{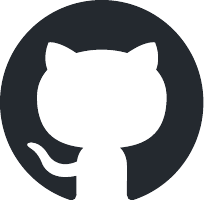} \href{https://github.com/GAIR-NLP/polymath-learning}{Code}
}
  
\maketitle

\thispagestyle{titlepage}

\begin{abstract}
The reasoning ability of large language models (LLMs) can be unleashed with reinforcement  learning (RL)~\citep{openai-o1, deepseekai2025deepseekr1incentivizingreasoningcapability, zeng2025simplerl}. The success of existing RL attempts in LLMs usually rely on high-quality samples of large volumes. In this paper, we challenge conventional assumptions about data requirements in RL for LLMs by demonstrating the effectiveness of one-shot reinforcement learning. Specifically, we introduce \textbf{\emph{polymath learning}}, a framework for designing one training sample that elicits multidisciplinary reasoning improvement. We present three key findings: (1) A single, strategically selected math reasoning sample can produce significant performance improvements across multiple domains, including physics, chemistry, and biology; (2) Analysis of salient mathematical skills provides insight into the characteristics associated with effective polymath samples; and (3) An engineered synthetic sample that integrates multidisciplinary elements and broader skill coverage achieves stronger performance than naturally occurring individual samples. Across various reasoning benchmarks, \textbf{\emph{polymath learning}} achieves stronger performance than larger datasets, demonstrating that reasoning structure and skills in samples, rather than quantity, may be the key to unlock enhanced reasoning capabilities in language models. 
Our results suggest a shift, dubbed as \textbf{\emph{sample engineering}}, toward precision engineering of samples that complements simply increasing data volume.
\end{abstract}

\pagestyle{normal} 

\section{Introduction}
Recent advances in Large Language Models (LLMs) have demonstrated the remarkable effectiveness of reinforcement learning (RL) in enhancing complex reasoning capabilities. Models like o1 \citep{openai-o1}, Deepseek R1 \citep{deepseekai2025deepseekr1incentivizingreasoningcapability}, and Kimi1.5~\citep{kimiteam2025kimik15scalingreinforcement} have shown that RL training is able to naturally induce sophisticated reasoning behaviors, including self-verification~\citep{weng-etal-2023-large}, reflection~\citep{shinn2023reflexion}, and extended chains of thought. While these advances typically rely on large-scale training data, recent work has begun to challenge this paradigm. \citet{li2025limrrlscaling} demonstrated with their LIMR approach that a strategically selected subset of just 1,389 samples can outperform the full 8k sample MATH dataset~\citep{hendrycks2021measuring}. More recently, \citet{wang2025reinforcement} made the surprising observation that even one single sample can produce meaningful improvements in math reasoning through RL, and \citet{wang2025unleashing} achieved similar gains by distilling high-quality reasoning paths from strong commercial models. However, this finding remains preliminary and math-specific, and leaves the critical questions of cross-domain generalization with internal abilities of LLMs unanswered: whether reasoning gains beyond math can be achieved in similar manner? Whether a strategy exists in directing the optimal sample? Whether such sample can be synthesized to enhance the sample quality?

In this paper, we build upon these emerging insights to investigate the phenomenon of one-shot reinforcement learning for broad reasoning tasks, termed as \textit{polymath learning}. Our central finding is that a single, carefully selected sample is able to produce significant performance gains not only in mathematics but across diverse domains including physics, chemistry, biology, as well as more general domains. This cross-domain generalization suggests that RL may enhance fundamental reasoning mechanisms rather than merely reinforcing domain-specific knowledge, even without saturated domain-specific training. Specifically, our work addresses 3 research questions:

\textbf{Cross-Domain Generalization in one-shot RL:} Does a single math sample yield improvements across diverse knowledge domains through polymath learning? We investigate the transfer mechanisms that allow reasoning patterns to transcend domain boundaries and observe that one single sample based on the math category selection elicits greater reasoning gains of LLMs than comprehensive datasets with thousands of samples. Notably, these gains extend beyond quantitative subjects to domains that are semantically distant from mathematics, highlighting the broad generalization potential of polymath learning.

\textbf{Salient Skill Identification:} What characteristics of polymath samples are associated with stronger cross-domain reasoning gains? Rather than relying on outcome-oriented difficulty metrics, we analyze the composition of salient math reasoning skills involved in solving candidate samples and find that samples exhibiting broader coverage of foundational mathematical skills tend to produce stronger multidisciplinary improvements. And the prevalence of algebra and precalculus skills across domains suggests their roles as broadly transferable reasoning components that facilitate multidisciplinary generalization.

\textbf{Synthetic Sample Construction:} How can we engineer a hybrid ``meta-sample" beyond naturally occurred ones that integrates multiple reasoning skills? We propose a synthesis technique through the lens of salient math skill taxonomy to construct the sample with comprehensive skill coverage and multidisciplinary context. The results illustrate that the multidisciplinary background strengthens the comprehensiveness of the salient skills, and yields greater cross-domain reasoning gains than the natural math samples that mainly possess math skills in limited categories and volume. Further analysis demonstrates that both the internal multidisciplinary reasoning structure and skills, as well as the computational complexity, contribute to reasoning power amplification of individual samples.

By demonstrating that a single sample can trigger broad and transferrable reasoning gains, our findings refine the current understanding of data requirements in RL, suggesting that the field may benefit from a shift toward ``\textbf{\emph{sample engineering}}'', that encourages deliberate selection, and synthesis of training samples to unlock reasoning capabilities more efficiently through high-quality data re-using~\citep{tan2025scalingbehaviorsllmreinforcement}, rather than simply scaling data volume, which may potentially induce generalization degradation~\citep{yang-etal-2024-unveiling, zhang2025generalizationmemorizationdynamicdecoding}.

\section{Related Work}
\paragraph{Reinforcement Learning in Language Models} Reinforcement learning has been applied to aligning language models with human intents~\citep{rlhf} or instructions~\citep{NEURIPS2022_b1efde53} through learning from human feedback. Later, it is extended to strengthen the long-form reasoning ability of models without relying on imitation of high-quality reasoning data, specifically by employing Reinforcement Learning with Verifiable Reward (RLVR) where the model outcomes can be verified and rewarded by verification functions with the advancement in RL algorithms~\citep{schulman2017proximalpolicyoptimizationalgorithms, lambert2025tulu3pushingfrontiers, hu2025reinforceefficientrlhfalgorithm}. However, training reliable outcome-based reward models~\citep{cobbe2021trainingverifierssolvemath} is challenging, and the rule-based reward function demonstrates effectiveness by simplifying the implementation of critic models and mitigating reward hacking~\citep{shao2024deepseekmath}. In this work, we extend the reasoning ability to broader domains by intensively learning from one training sample that is selected to carry high quality reasoning structure.

\paragraph{Data Efficiency in Reinforcement Learning}
\citet{xu2025rolloutsusefuldownsamplingrollouts} selects variance-based subset responses for GRPO training. \citet{zhang2025srpocrossdomainimplementationlargescale} employs the most recent reward information for filtering prompts, which is beneficial to GRPO training~\cite{yu2025dapoopensourcellmreinforcement}. Other than focusing on the response quality in RL training, \citet{li2025limrrlscaling} highlights the significance of prompt quality by demonstrating the effectiveness of carefully selected training subset. Further, \citet{shrestha2025warmtrainunlockinggeneral} demonstrates cross-domain reasoning ability with less than 100 samples but requires a pre-warmup distillation stage, and \citet{wang2025reinforcement} utilizes only one training sample and achieves a notable improvement in mathematical reasoning. And \citet{zhao2025absolutezeroreinforcedselfplay} requires no human-expert data but still relies on an external executor to generate valid answers to synthetic coding problems. However, these studies still focusing on the mathematical reasoning domain where the training data originates and neglect its broader impacts on multiple disciplines where the reasoning ability essences.

\paragraph{Transfer Learning and Cross-Domain Generalization}
\citet{afzal-etal-2024-adapteval} demonstrates that small LLMs can catch up with larger counterparts in domain adaptation with few examples. And ~\citet{chen-etal-2024-style} adapts models to new domain by extracting domain-invariant features in existing domain. For reasoning problems, \citet{zhao2025absolutezeroreinforcedselfplay} unleashes an improvement in mathematical reasoning soly based on training on programming data, and ~\citet{huan2025doesmathreasoningimprove} demonstrates that RL achieves better generalization from math to other domains than supervised fine-tuning, without a deep dive into data efficiency. \citet{li2025domainhelpothersdatacentric} investigates the cross-domain impact from math reasoning, but only limits the study within logical-intensive domains like code and puzzle. In polymath learning, we enlarge the reasoning scope to various subjects and study the learning impact from one sample.

\paragraph{Sample Selection Strategies}
The effectiveness of finetuning large language models is heavily dependent on the quality of data selection~\citep{xie2023dataselectionlanguagemodels}. And well selected data samples can unlock powerful fine-tuning performance compared to larger data volumes~\citep{wang-etal-2023-self-instruct, NEURIPS2023_ac662d74}. \citet{xia2024lessselectinginfluentialdata} relies on gradient information for data selection, while \citet{tsds} formulates data selection as an optimal transportation problem. The effectiveness of data selection also extends to reasoning problems~\citep{qin2024o1, ye2025limoreasoning}. Various approaches are introduced in sample selection in RL training of LLMs, including LLM-based scores or justification~\citep{liu2024what, li2025addoneinincrementalsampleselection}, clustering-based utility~\citep{yang2026adaptive}, sample difficulty~\citep{havrilla2025sparqsyntheticproblemgeneration, sun2025improving, Polaris2025}, LLM-based role-play~\citep{luo2025personamathboostingmathematicalreasoning} and gradient alignment~\citep{yang2026gradaligngradientaligneddataselection} to estimate sample diversity for data selection. In polymath learning, we select polymath samples based on the alignment with reinforcement learning dynamics and salient math skills to unlock the reasoning ability in multiple disciplines.

\section{GRPO Basics}
\label{sec:GRPO basics}
Given a dataset $\mathcal{D}=\{(x, \hat{y})\}$ where $x$ and $\hat{y}$ stand for the prompt and golden answer, RLVR relies on a policy model $\pi_{\theta}(\cdot|x)$ to generate correct reasoning trajectories without relying on trajectories generated by human-expert or teacher models~\citep{zhao2025absolutezeroreinforcedselfplay}. In GRPO~\citep{shao2024deepseekmath}, the advantage value is estimated within a group of responses  $G$ responses $\{y_{1}, y_{2}, ..., y_{G}\}$ to substitute the critic model in PPO while remaining effectiveness. Specifically,
\begin{align*}
\mathcal{L}_{GRPO} & = \mathrm{E}_{[x\sim\mathcal{D}, \{y_{i}\}\sim\pi_{\theta_{\text{old}}}(\cdot|x)]}[\frac{1}{G}\sum_{i=1}^{G}\frac{1}{|y_{i}|}\sum_{t=1}^{|y_{i}|}\text{min}(\tilde{r}_{i,t}A_{i}, \text{clip}(\tilde{r}_{i,t}, 1-\epsilon, 1+\epsilon)A_{i})-\beta KL(\pi_{\theta}||\pi_{ref})]\\
A_{i} & = \frac{r_{i} - \text{mean}(r_{1}, r_{2}, ... r_{G})}{\text{std}(r_{1}, r_{2}, ... r_{G})}, \quad \tilde{r}_{i,t} = \frac{\pi_{\theta}(y_{i,t}|x, y_{i, <t})}{\pi_{\theta_{\text{old}}}(y_{i,t}|x, y_{i, <t})}
\end{align*}
Here $r_{i}$ is computed by applying the reward function on the response and the golden answer $r_{i} = \text{reward}(y_{i}, \hat{y}_{i})$. $\pi_{\theta}(y_{i,t}|x, y_{i, <t})$ identifies the likelihood of the $t$-th token in $i$-th response from the policy model. Unlike previous efforts that assembles $\mathcal{D}$ with a comprehensive set of samples, in polymath learning, the dasataset consists of one valid sample and $\mathcal{D}_{polymath} = (x_{1}, \hat{y}_{1})$. 

\section{Polymath Learning}
~\citet{openai2024openaio1card} unlocks complex reasoning ability of LLM through reinforcement learning, and ~\citet{deepseekai2025deepseekv3technicalreport, deepseekai2025deepseekr1incentivizingreasoningcapability} further demonstrates that such advanced reasoning ability can be elicited directly from pretrained base models using rule-based rewards, without relying on imitation from high-quality supervised reasoning trajectories. Existing explorations mainly focus on math or synthetic logic~\citep{zeng2025simplerl, tinyzero, xie2025logicrlunleashingllmreasoning} where large volumes of questions with rule-based verifiable answers are accessible. Beyond the success of \textit{comprehensive learning}: training models with thousands of comprehensive high-quality problems and beyond, \citet{wang2025reinforcement} shows that the reasoning ability can also be boosted by one single math sample with RL. Following this inquiry, we investigate \textit{polymath learning}: training with one sample that plays a polymath role and extends the model reasoning power across domains. Similar to \citet{wang2025reinforcement}, we conduct polymath learning from math reasoning problems.

\paragraph{Polymath Learning with One Natural Sample}
LIMR~\citep{li2025limrrlscaling} displays the potential of improving training efficiency in reinforcement learning by selecting a subset of samples from MATH that closely align with the training dynamics of RL. A preliminary model is trained in LIMR to record the reward trajectories during optimization. The sample learnability is then computed by comparing its outcome reward with the dataset-wise average of outcome rewards. Higher LIMR scores indicate greater alignment between the model behavior on individual sample and the entire dataset during RL training. However, learning from samples with excessively high LIMR scores risks over-specialization in math reasoning at the expense of the broader reasoning capabilities in other disciplines. Therefore, we select LIMR samples with the lowest scores (0.6) in different math categories as polymath candidates to maintain the same learnability according to preliminary experiments. One polymath sample is displayed in Table~\ref{tab: algebra-math-polymath learning sample} and others are included in Appendix~\ref{sec: other polymath samples}.

\begin{table*}[ht!]
\begin{tcolorbox}[colback=blue!5!white,colframe=black,width=1.0\textwidth,title={Polymath Sample in Algebra}]
\small
\examplequestion{[Question]}
A $100$-gon $P_1$ is drawn in the Cartesian plane. The sum of the $x$-coordinates of the $100$ vertices equals 2009. The midpoints of the sides of $P_1$ form a second $100$-gon, $P_2$. Finally, the midpoints of the sides of $P_2$ form a third $100$-gon, $P_3$. Find the sum of the $x$-coordinates of the vertices of $P_3$. \\

\examplequestion{[Answer]}
\( \boxed{2009} \)
\end{tcolorbox}
\vspace{-0.2cm}
\caption{Polymath sample in algebra.}
\label{tab: algebra-math-polymath learning sample}
\end{table*}

\paragraph{Polymath Learning with One Synthetic Sample}
Synthesizing reasoning trajectories have been shown beneficial in boosting the reasoning ability in LLM in the pretraining~\citep{ishibashi2025mininghiddenthoughtstexts} and supervised-finetuning stage~\citep{singh2024beyond, yuan2024scaling}. Careful problem synthesis also scales up the mathematical reasoning ability of models by reinforcement learning~\citep{setlur2024rl}. Since solving multidisciplinary problems and purely mathematical problems are not require on the same base of expertise, existing problem synthesis approaches based on problem imitation~\citep{toshniwal2025openmathinstruct}, mutation~\citep{havrilla2025sparqsyntheticproblemgeneration} or creation based on seed concept or problem bank~\citep{key_point_synthesis_Huang_Liu_Gong_Gou_Shen_Duan_Chen_2025, liang2025swsselfawareweaknessdrivenproblem, zhao-etal-2025-promptcot, liu2025designerdesignlogicguidedmultidisciplinarydata} do not directly apply. In practice, we find it challenging to organically integrate and align information from problems in diverse disciplines. Therefore, unlike \citet{setlur2024rl} and \citet{wang2025unleashing}, we synthesize the polymath sample based on instruction without relying on existing problems or models finetuned with question-generation~\citep{ding2025unleashingllmreasoningcapability, wu2025synthrlscalingvisualreasoning}. Our final problem synthesis pipeline includes two stages,
\begin{itemize}
    \item \textbf{Candidate problem generation} We employ strong models like OpenAI-O3~\citep{openai2025o3}, Gemini2.5-Pro~\citep{google2025gemini25pro} and DeepSeek-R1 to integrate knowledge from physics, chemistry, and biology. The golden answers are collected from the joint success in problem solving of these models.
    \item \textbf{Specialized problem selection} After massive collection of candidate problems, we employ Qwen2.5-72B-instruct to identify the salient math skills related in solving the problem given the problem text. The abundance of skills in different math categories is employed to reflect the complexities and qualities of problems. We then select the problems with the most specialized skills as the synthesized polymath samples, please refer to Appendix~\ref{sec: other config} for the prompt employed and Appendix~\ref{sec: math skill statistics} for example.
\end{itemize}

We find this instruction-based approach unleashes the creativity of LLMs in producing complex multidisciplinary problems. Specifically, we select the synthesized polymath sample with the most comprehensive skill spectrum (\syntheticprimenospace, shown in Table~\ref{tab: polymath-shot-synthetic}). Solving the \syntheticprime requires a complex set of knowledge, including the strand sequence (biology), chemical bonds and energy to break bonds (chemistry), accumulating energy by collecting photons and estimating photon energy based on its wavelength (physics). The synthesis prompt is shown in Appendix~\ref{sec: other config}.

\begin{table*}[ht!]
\begin{tcolorbox}[colback=blue!5!white,colframe=black,width=1.0\textwidth,title={Polymath Sample (Synthetic Prime)}]
\small
\examplequestion{[Question]}
A double‑stranded DNA fragment of exactly 11 base pairs has the upper strand sequence\\
5 'G C G C G C G C A T A 3'.\\

Each adenine–thymine (A·T) base pair is held together by **2** hydrogen bonds, and each guanine–cytosine (G·C) base pair by **3** hydrogen bonds.

The DNA molecule is irradiated with monochromatic light of wavelength $\lambda = 400 \text{nm}$.
Assume that **100\%** of every photon’s energy is used exclusively to break hydrogen bonds between the two strands.

Use the exact data below (treat every value as exact):

* Enthalpy of one hydrogen bond $\Delta H = 20 \text{kJ · mol}^{-1}$\\  
* Planck constant $h = 6.626 \times 10^{-34} \text{J · s}$ \\
* Speed of light $c = 3.00 \times 10^{8} \text{m· s}^{-1}$ \\ 
* Avogadro constant $N_{a} = 6.022 \times 10^{23} \text{mol}^{-1}$ \\

**Fill in the blank:**  
What is the minimum number of 400 nm photons required to supply exactly enough energy to dissociate *all* hydrogen bonds in **one** molecule of this DNA fragment?  
(Answer with a single positive integer.)\\

\examplequestion{[Answer]}
\( \boxed{2} \)
\end{tcolorbox}
\vspace{-0.3cm}
\caption{The synthetic prime polymath sample that incorporates multidisciplinary knowledge.}
\label{tab: polymath-shot-synthetic}
\end{table*}

\section{Experimental Setup}
\label{sec: experimental setup}
We choose Qwen2.5-7b-base~\citep{qwen2025qwen25technicalreport} as the primary model due to its balance in general reasoning ability and exploration space to observe isolatable performance change. Similar to \citet{wang2025reinforcement}, we adopt RLVR by employing GRPO~\citep{shao2024deepseekmath} for training. The selected polymath sample is augmented into the batch of 128. For each prompt, we generate 16 responses with temperature of 1.0 to collect diverse reasoning trajectories. The prompt template follows the design of~\citet{hu2025openreasonerzeroopensourceapproach}. Following~\citet{huan2025doesmathreasoningimprove}, the model is trained for 140 steps since the reasoning ability saturates. We employ a simple binary outcome reward with rule-based matching of the final answer according to its effectiveness in previous studies~\citep{shao2024deepseekmath, yu2025dapoopensourcellmreinforcement}, and exclude the format reward and the KL term as they demonstrate inferior performance~\citep{wang2025reinforcement, yu2025dapoopensourcellmreinforcement}. In skill identification, we employ \textit{Algebra} to include salient skills from \textit{Prealgebra}, \textit{Algebra} and \textit{Intermediate Algebra} to eliminate their large overlaps. We discuss practices in additional model choices in Appendix~\ref{sec: additional model polymath learning} and RL method in Appendix~\ref{sec: polymath learning other RL}. 

To evaluate cross-domain generalization, we construct a diverse benchmark suite covering both mathematical and non-mathematical reasoning domains. Specifically, we select MATH500~\citep{lightman2024lets}, AIME in 2024 and 2025, MinervaMath~\citep{minervamath}, GPQA-Diamond~\citep{rein2024gpqa}, Scibench~\citep{wang2024scibench}, MMLU-Pro~\citep{wang2024mmlupro} with randomly selecting 100 problems for each subject and SuperGPQA~\citep{pteam2025supergpqascalingllmevaluation}  with 1500 random problems as the evaluation set. The full spectrum of subjects is listed in Appendix~\ref{sec: full eval subject list}. The model responses are generated with greedy decoding in single attempt, except for AIME, where the results are averaged from 32 attempts with temperature being 0.4. We include additional configurations in  Appendix~\ref{sec: other config}.

\begin{table}[h]
\centering
\resizebox{1.0\textwidth}{!}{
\begin{tabular}{lccccccccc}
\toprule
\textbf{Data} & \textbf{Math} & \textbf{Physics} & \textbf{Chemistry} & \textbf{Biology} & \textbf{Science} & \textbf{Engineering} & \textbf{Computer Science} & \textbf{Others} &  \textbf{Avg} \\
\rowcolor{lightblue}
\multicolumn{10}{c}{\textbf{N=64 Sampling (0-shot)}}\\
0-shot & 20.4 & 4.4 & 4.4 & 5.1 & 0.0 & 3.7 & 3.3 & 9.6 & 6.4 \\
\rowcolor{lightblue}
\multicolumn{10}{c}{\textbf{In-context Learning (1-shot)}}\\
\rowcolor{SampleSeparator}
\multicolumn{10}{c}{\textbf{Natural Sample}}\\
Geometry & 24.5 & 8.0 & 7.2 & 24.4 & 4.3 & 6.0 & 29.0 & 11.6 & 14.4 \\
\midrule
Prealgebra & 22.3 & 11.2 & 9.4 & 40.3 & 6.8 & 10.2 & 35.0 & 20.3 & 19.4 \\
\midrule
Algebra & 21.4 & 10.9 & 9.8 & 38.7 & 8.3 & 10.4 & 35.0 & 20.6 & 19.4 \\
\midrule
Intermediate Algebra & 22.7 & 8.0 & 7.0 & 21.8 & 4.5 & 9.5 & 32.0 & 15.5 & 15.1 \\
\midrule
Number Theory & 21.7 & 10.9 & 8.7 & 31.9 & 5.4 & 6.6 & 28.0 & 15.8 & 16.1 \\
\midrule
Precalculus & 21.6 & 8.3 & 5.9 & 20.2 & 5.2 & 6.8 & 26.0 & 11.9 & 13.2 \\
\midrule
Probability & 22.4 & 9.7 & 7.2 & 24.4 & 5.6 & 7.7 & 22.0 & 13.2 & 14.0 \\
\rowcolor{SampleSeparator}
\multicolumn{10}{c}{\textbf{Synthetic Sample}}\\
Prime & 18.6 & 4.6 & 4.6 & 8.4 & 2.2 & 4.6 & 11.0 & 7.7 & 7.7 \\
\rowcolor{lightblue}
\multicolumn{10}{c}{\textbf{Comprehensive Learning ($>$ 1k shots)}}\\
\rowcolor{SampleSeparator}
\multicolumn{10}{c}{\textbf{Natural Sample}}\\
MATH & 37.2 & 12.8 & 10.0 & 31.4 & 6.5 & 8.6 & 25.8 & 23.4 & 19.5 \\
\midrule
LIMR & 38.0 & 11.6 & 11.8 & 48.3 & 10.0 & 13.4 & 35.1 & 31.5 & 25.0 \\
\rowcolor{lightblue}
\multicolumn{10}{c}{\textbf{Polymath Learning (1-shot) - Ours}}\\
\rowcolor{SampleSeparator}
\multicolumn{10}{c}{\textbf{Natural Sample}}\\
Geometry & 15.5 & 9.9 & 10.0 & \textbf{55.1} & 11.2 & 16.7 & 37.1 & \textbf{35.0} & 23.8 \\
\midrule
Prealgebra & 38.0 & 17.4 & 12.2 & 51.7 & 15.1 & 16.5 & \textbf{49.5} & 33.5 & 29.2 \\
\midrule
Algebra & 37.3 & 17.4 & 13.7 & 51.7 & 12.1 & 15.6 & 43.3 & 30.9 & 27.7 \\
\midrule
Intermediate Algebra & 36.3 & 19.1 & 13.1 & 50.0 & 13.9 & 17.5 & 42.3 & 31.1 & 27.9 \\
\midrule
Number Theory & 37.7 & 16.9 & 12.4 & 49.2 & 13.4 & 17.8 & 42.3 & 32.2 & 27.7 \\
\midrule
Precalculus & 38.0 & 18.4 & 13.7 & 50.0 & 16.0 & 19.7 & 43.3 & 31.0 & 28.8 \\
\midrule
Probability & \textbf{38.8} & 19.9 & 11.5 & 46.6 & 14.7 & 16.4 & 41.2 & 31.4 & 27.6 \\
\rowcolor{SampleSeparator}
\multicolumn{10}{c}{\textbf{Synthetic Sample}}\\
Geometry & 36.2 & 15.0 & 11.5 & 52.5 & 11.0 & 13.2 & 36.1 & 31.1 & 25.8 \\
\midrule
Algebra & 37.6 & 16.9 & 12.6 & 52.5 & 13.9 & 18.6 & 41.2 & 31.5 & 28.1 \\
\midrule
Number Theory & 39.4 & 18.2 & 12.0 & 47.5 & 13.8 & 18.6 & 36.1 & 32.1 & 27.2 \\
\midrule
Precalculus & 37.6 & 20.3 & 15.3 & 48.3 & \textbf{16.5} & \textbf{20.8} & 44.3 & 32.9 & 29.5 \\
\midrule
Probability & 37.2 & 16.7 & 13.9 & 50.0 & 10.8 & 19.7 & 46.4 & 30.1 & 28.1 \\
\midrule
Prime & 38.3 & \textbf{20.6} & \textbf{15.7} & 54.2 & 15.6 & \textbf{20.8} & 48.5 & 32.4 & \textbf{30.8} \\
\bottomrule
\end{tabular}}
\caption{The performance of employing different sample strategies on different subject domains. The best performance on each subject domain is \textbf{bolded}. Most natural polymath samples outperforms in-context learning and comprehensive learning with LIMR selection. Most synthetic specialist samples outperforms the corresponding natural sample, and the \syntheticprime sample demonstrates the best performance. The dataset-wise results and robustness experiments are included in Appendix~\ref{sec: results by datasets} and Appendix~\ref{sec: result robustness} respectively.}
\label{table: main results by subject domains}
\end{table}

\section{Results}
\subsection{Cross-Domain Generalization of Learning on Single Polymath Sample}
Table~\ref{table: main results by subject domains} reports the reasoning performance aggregated by subject domains (e.g. \textit{Math} includes all math problems from MATH500, AIME, MinervaMath and other benchmarks). Models trained with various natural and synthetic polymath samples are compared against the base model. Other than the \syntheticprime sample, we construct several synthetic specialist samples across different math categories by selecting instances containing the highest number of salient skills identified in those categories. Here, we make several observations. Firstly, the base model exhibits imbalanced reasoning abilities: performing strongly in math but weakly in other domains. Secondly, polymath learning delivers substantial improvements over in-context learning across different subject domains. Thirdly, although comprehensive learning enhances the math reasoning ability of the base model, especially with effective data selection strategies like LIMR, most natural polymath samples demonstrate comparable performance to comprehensive learning on the math domain, and surpass it on non-math domains, underscoring the potential of single high-quality sample in unlocking reasoning ability. Notably, polymath samples in prealgebra and precalculus stand out, exhibiting superior performance due to their abundant salient skills (Sec~\ref{sec: characteristics of optimal polymath sample}). Lastly, synthetic polymath samples further elevate the reasoning ability. Most specialist samples outperform their natural polymath sample counterparts and demonstrate domain specific advantages: geometry and algebra samples excel in engineering, and probability sample in science. Furthermore, the \syntheticprime sample achieves the strongest overall performance and demonstrates particular strength in physics and chemistry, suggesting that the reasoning potential of individual samples can be amplified through well-incorporation of multidisciplinary knowledge. Therefore we select the \syntheticprime sample as the primary synthetic sample for subsequent experiments. Unlike data collection approaches that are based on widely crawled sources~\citep{wu2025reasoningmemorizationunreliableresults, he2025deepmath103klargescalechallengingdecontaminated, zhang2025largescalediversesynthesismidtraining}, our polymath samples do not rely on seed data to construct or displaying evidence of data contamination. Please refer to Appendix~\ref{sec: other polymath samples} for the specialist samples. We further factorize the effectiveness of the \syntheticprime sample in Appendix~\ref{sec: synthetic prime variants}.

\begin{figure}[h]
    \centering
    \includegraphics[width=0.99\linewidth]{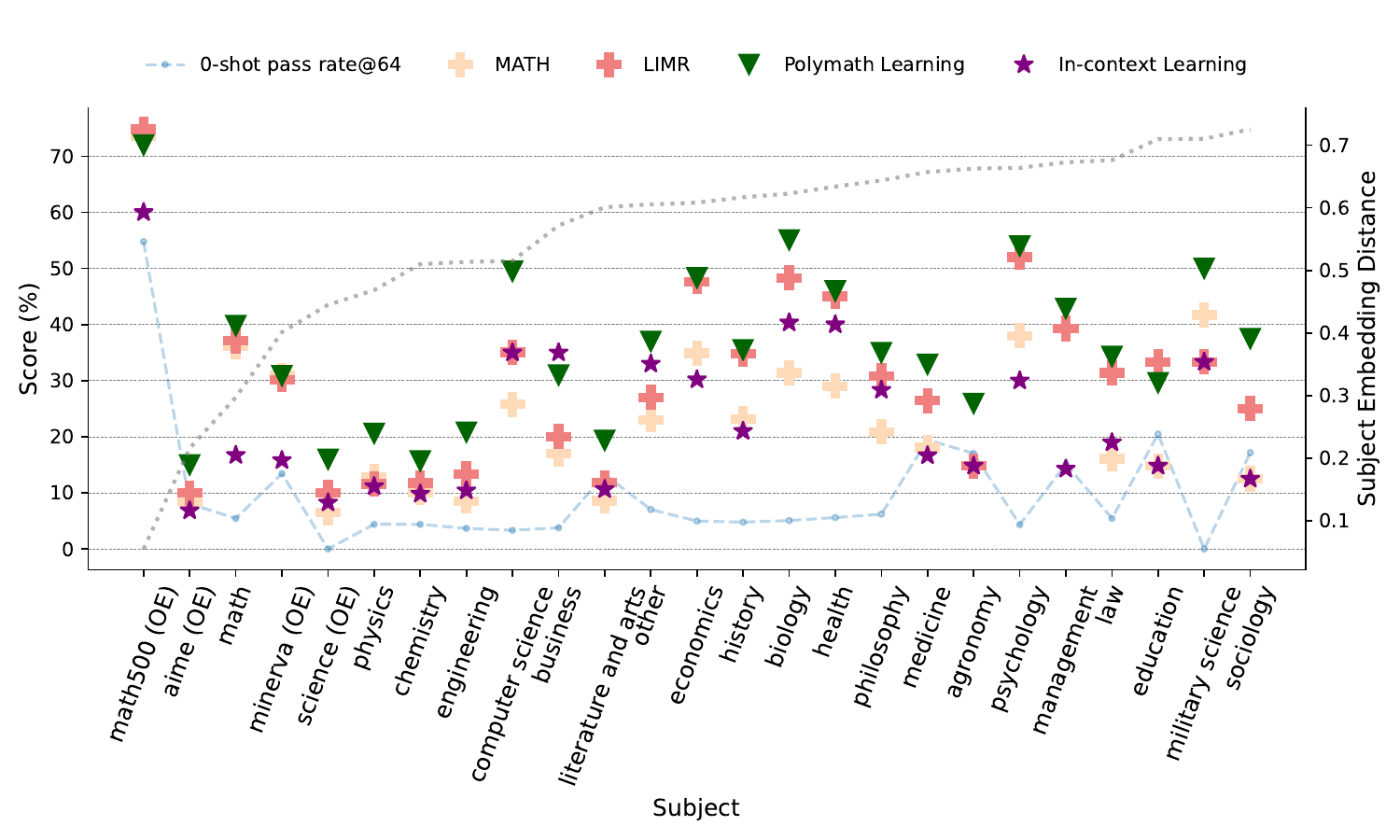}
    \vspace{-0.2cm}
    \caption{The subject-level performance of different learning strategies. \textit{OE} stands for subjects with open-ended problems. The subjects are sorted by subject embedding distance to MATH500 (the grey dotted line), from low to high. The blue line represents pass ratio from 64 independent attempts of the base model. The stars and triangles represent best performance of in-context learning and polymath learning. Note that we only display the best polymath learning and in-context polymath learning results for demonstration.}
    \label{fig:polymath learning perf by subject embedding}
\end{figure}

The breakdown performance of N sampling (0-shot pass rate@64), polymath learning and comprehensive learning by subjects is visualized in Figure~\ref{fig:polymath learning perf by subject embedding}, with subjects ordered by their similarities to math. The similarity is measured by computing the subject embedding distance between the average embedding of all problems in each subject and the mean of problems in MATH500. We employ Text-Embedding-3-Small~\citep{openaitextembedding3small} with dimension of 1024 to generate problem representations. The best performance of polymath learning and in-context learning of polymath samples are displayed with triangles and stars, respectively. We include our major findings,

\paragraph{Strong mathematical but skewed reasoning of the base model} Due to the massive mathematical and coding data participated in pretraining~\citep{qwen2025qwen25technicalreport, wu2025reasoningmemorizationunreliableresults}, the Qwen2.5-7b-base model achieves $\text{pass rate@64} > 0.5$ in MATH500, higher than all other subjects with significant margins. However, the strength in MATH500 does not naturally extend to other subjects. For example, the base model performs poorly on physics, chemistry and biology, but demonstrates relative strength ($\text{pass rate@64}$ close to 0.2) in education, medicine, sociology and management, which does not possess similar proportion of quantitative components as math does.

\paragraph{Comprehensive learning provides mathematical dominance, but not multidisciplinary} Comprehensive learning with MATH or LIMR sets demonstrate strong performance in MATH500, and remain competitive with the strongest polymath sample in other math subjects (math, minerva). However, their performance on most non-math subjects lags far behind from the best polymath results. The reasoning strengths gained from math-specific training generalize only to a limited set of subjects, like economics, health, psychology, education, and history where more than fourfold performance improvement over zero-shot reasoning is observed. Nonetheless, quality-driven data selection remains beneficial in comprehensive learning, with LIMR consistently outperforming MATH in most subjects. The training dynamics further reveals that even though comprehensive learning excels in in-domain math evaluation, it suffers from overfitting in multidisciplinary benchmarks where polymath learning provides more stable performance (see Appendix~\ref{sec: training dynamics} for details).

\paragraph{The effectiveness of in-context learning of polymath samples}The best in-context polymath learning sample outperforms the  0-shot pass rate@64 baseline in most subjects, highlighting the efficacy of polymath samples even under gradient-free learning. Moreover, we observe that the specific polymath samples (e.g. prealgebra or algebra) are able to achieve performance on par with, or superior to, at least one model trained via comprehensive learning in over 50\% of subjects, with details included in Appendix~\ref{sec: reasoning breakdown by subject}.

\paragraph{Better generalization of polymath learning on math-distant subjects} Even though the best polymath sample outperforms comprehensive learning in LIMR on math-intensive domains like math and engineering, its advantage is more pronounced on subjects that are semantically distant from math. For example, it demonstrates around 10 points gains in agronomy, literature and sociology. On average, polymath learning with the best natural samples yields a 14.5 points improvement over comprehensive learning on the full MATH set on the 50\% subjects farthest from MATH500, compared to a 7.7 points gain on the 50\% subjects closest to MATH500. This pattern suggests that polymath learning promotes stronger reasoning generalization in less math-intensive subjects.

\subsection{Characteristics of Polymath Sample in Salient Math Skills}
\label{sec: characteristics of optimal polymath sample}
\begin{figure}[h]
    \centering
    \includegraphics[width=1.0\linewidth]{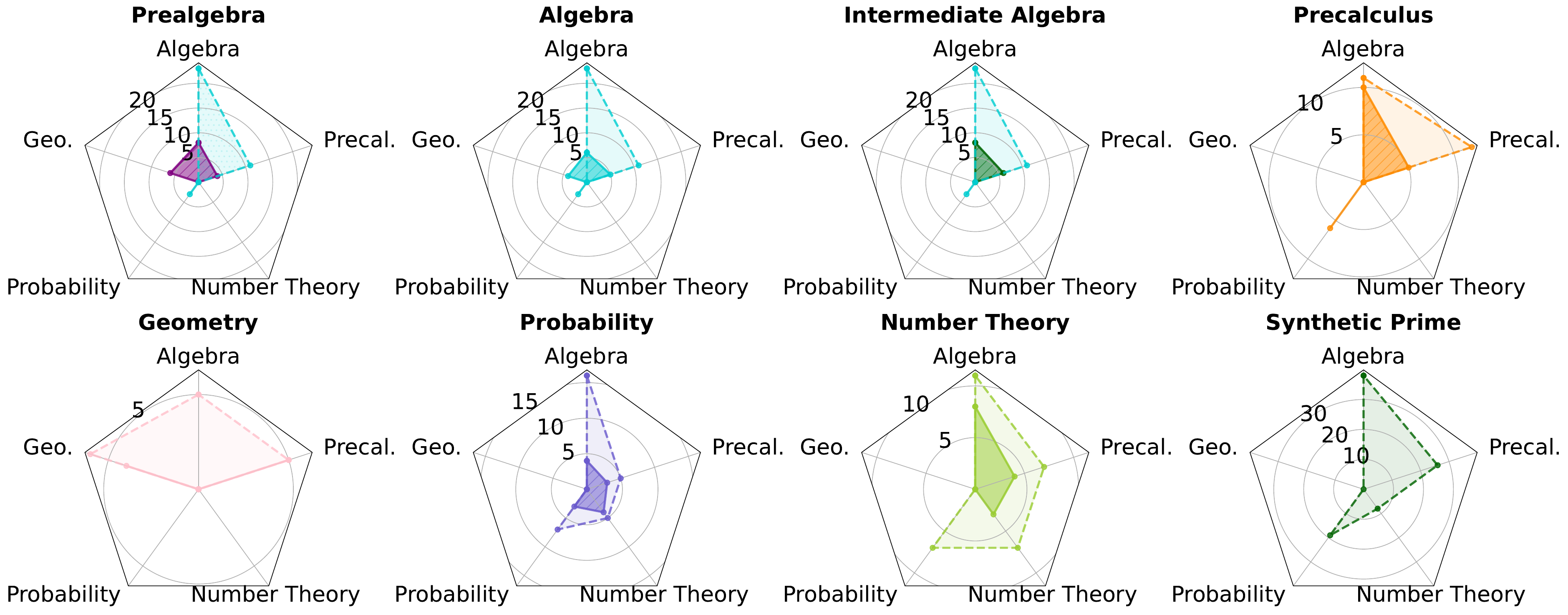}
    \caption{Skill spectrum between natural and synthetic polymath samples. The polygon represents number of salient skills identified in each math domain (\textit{Geo.} and \textit{Precal.} represents \textit{Geometry} and \textit{Precalculus} respectively). The real and dashed areas represent the natural and synthetic specialist samples except the last one, which represents the \syntheticprime sample, and the synthetic samples include more comprehensive salient skill sets than the natural polymath samples.}
    \label{fig:natural and synthetic math skill radar}
\end{figure}

Data diversity is beneficial in training more capable reasoning LLMs~\citep{zhang2025largescalediversesynthesismidtraining}, serving both regularization to the neural network~\citep{ba2025datadiversityimplicitregularization} and a mean to mitigate performance saturation especially when leveraging synthetic data sources~\citep{prismatic-synthesis}. In polymath learning, we extend beyond the diversity at the level of problem or trajectory~\citep{yu2025flowreasoningtrainingllms} and instead examine the composition of salient mathematical skills within individual polymath samples. The result in Figure~\ref{fig:natural and synthetic math skill radar} illustrates the key supporting role of algebra and precalculus skills in cross-domain reasoning. Polymath samples demonstrate stronger performance tend to exhibit high prevalence of these skills. Furthermore, synthetic specialist samples with multidisciplinary context span a broader range of skills than natural math samples specialized in the same category, which accounts for their superior performance. Notably, the \syntheticprime sample exhibits the highest concentration of salient skills, suggesting that solving such problems requires a complex interplay of knowledge and thus provides rich learning signals for training. The comparison with other out-of-MATH 1-shot sample is included in Appendix~\ref{sec: pi_1} where the \syntheticprime sample stays effective.

\begin{figure}[h]
    \centering
    \includegraphics[width=1.0\linewidth]{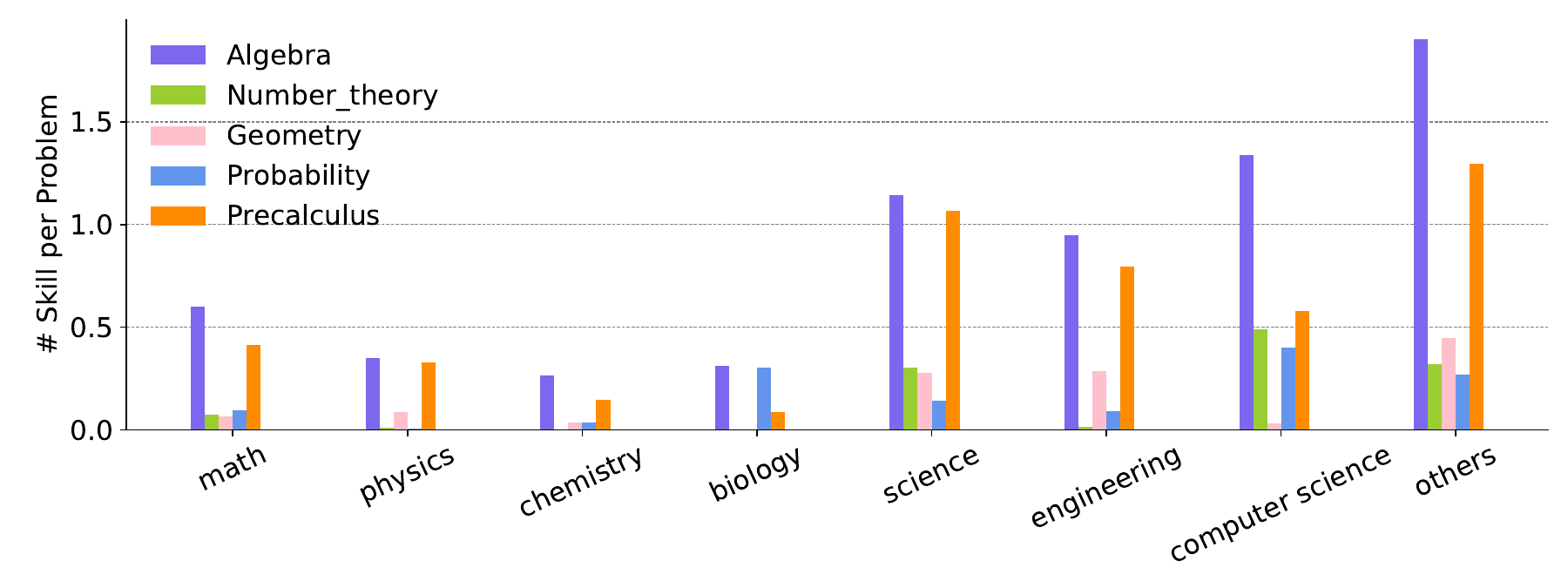}
    \caption{Average number of mathematical skills employed per problem in different subject domains. \textit{Algebra} and \textit{Precalculus} skills are the most prevalent.}
    \label{fig:math concept statistics}
\end{figure}

The distribution of salient skills across subject domains further highlights the central roles of algebra and precalculus. Skill abundance also reflects the degree of domain specialization. For instance, in engineering, the most frequent algebraic and geometric skills are \textit{unit conversion} and \textit{trigonometry}. Figure~\ref{fig:math concept statistics} shows that algebra and precalculus consistently dominate in skill popularity, underscoring their foundational importance for quantitative reasoning (e.g., \textit{unit conversion} and \textit{arithmetic operations}). Moreover, domains with integrative knowledge, such as science and engineering, demand more comprehensive combinations of salient skills compared to discipline-focused domains such as physics, chemistry, or biology.

\section{Generalization of Self-Verification}
\label{sec: self-verification}
The verification mechanism act as a signal for models to reconsider and refine their initial solutions~\citep{deepseekai2025deepseekr1incentivizingreasoningcapability}. Verification feedback can further enhance decision-making~\citep{madaan2023selfrefine, shinn2023reflexion}. To analyze such behavior, several signature words have been proposed for monitoring self-verification patterns~\citep{xie2025logicrlunleashingllmreasoning}. Following this, we collect pattern statistics across polymath learning samples, adding the `code' category to capture python-based program verification and excluding `reevaluate' for its rare appearance. We find that polymath learning in general demonstrates more frequent self-verification behavior than comprehensive learning. Moreover, the polymath sample in `number theory' and `intermediate algebra' exhibit strong tendencies in eliciting the self-checking (`re-evaluate') behavior and programming assistance (`code') respectively. Moreover, different polymath samples display distinct self-verification preferences depending on the subject domain, with details in Appendix~\ref{sec: self-verification by sbuject domain}. 

\begin{figure}[h]
    \centering
    \includegraphics[width=0.99\linewidth]{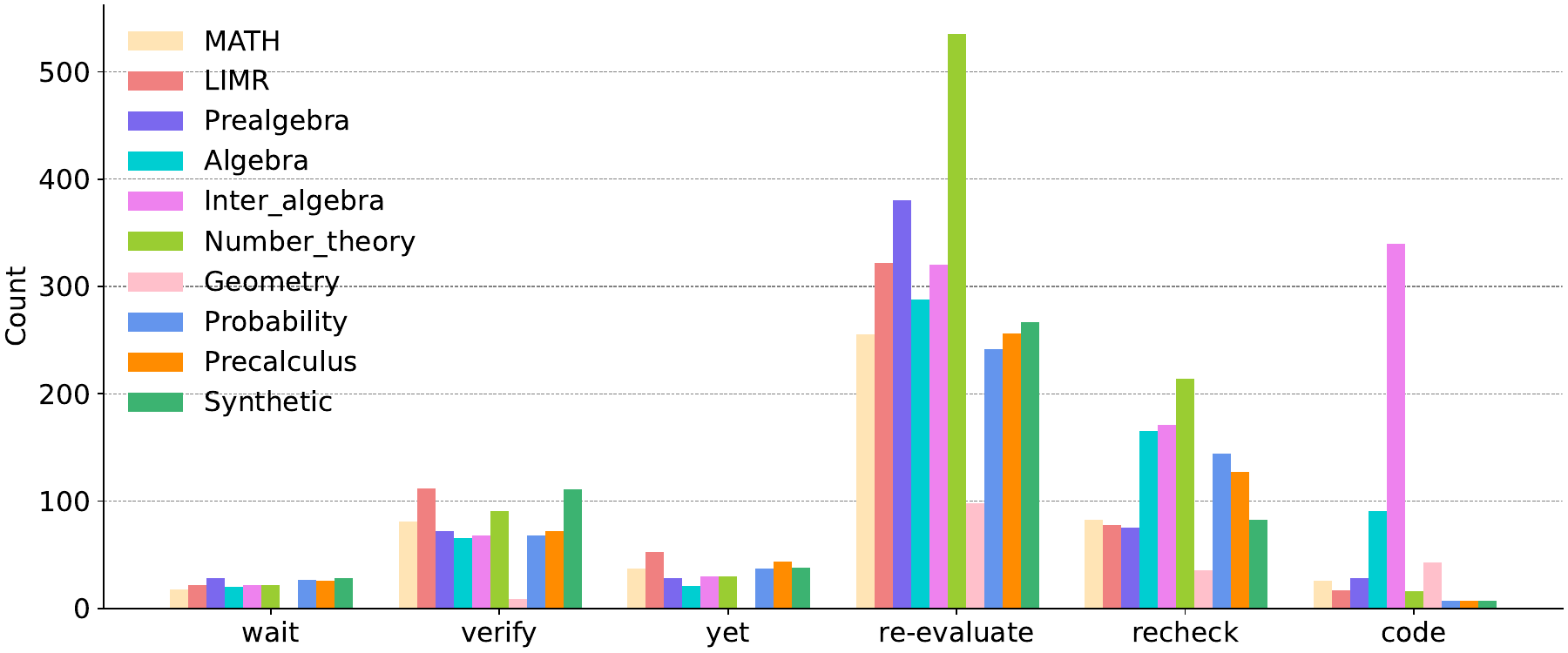}
    \caption{Self-verification patterns under different comprehensive and polymath samples across all subjects. Verification patterns like `re-evaluate' and `recheck' appear most frequently in polymath learning with the `number theory' sample, and the `intermediate algebra' sample elicits the most code blocks in reasoning.}
    \label{fig:self verification words}
\end{figure}

Similar to \citet{shao2025spuriousrewardsrethinkingtraining}, we observe frequent use of program verification in the polymath sample of `intermediate algebra'. However, the role of programs varies across domains: the programs in math are primarily used as part of the final answer generation process, including pseudo-execution errors like `Timed out'; in physics and chemistry, by contrast, the programs are employed more for result validation. Importantly, without the access of external executor, the integration of program does not necessarily yield reasoning gains. Illustrative examples are provided in Appendix~\ref{sec: self-verificiation example}.

\section{Conclusion}
While math reasoning ability has been regarded the primary metric for measuring progress in LLM reasoning, the broader multidisciplinary reasoning abilities remain relatively underexplored. Inspired by the success of boosting math reasoning ability with one single math training sample, we introduce polymath learning, a paradigm demonstrating that RL with a single carefully designed training sample can significantly improve reasoning ability across multiple disciplines beyond math. This approach rivals or even surpasses datasets by orders of magnitude in unlocking reasoning improvements across diverse domains and model choices. 
We further identify the coverage of salient mathematical skills as the key characteristic of effective polymath samples. Building on these insights, we propose a synthetic sample construction method that integrates multidisciplinary knowledge with comprehensive skill coverage, achieving stronger performance than naturally occurring problems. These results highlight the critical role of sample structure and reasoning richness in driving reasoning improvements.
Our findings challenge the prevailing assumption that scaling data volume is necessary for advancing reasoning ability. They instead point toward a complementary sample engineering direction, where carefully designed training examples provide dense learning signals that unlock broad generalization in reasoning.  

\bibliographystyle{acl_natbib}
\bibliography{related}

\appendix

\clearpage

\section{Limitations and Future Work}
In polymath learning, we focus our study in the effectiveness of one single training sample in lifting interdisciplinary reasoning ability with reinforcement learning. Due to resource constraints, our study only covers a small set of samples without larger-scale experiments in one-shot polymath learning. And the sample selection based on salient skills does not extend to scaled skill-based problem synthesis like \citet{havrilla2025sparqsyntheticproblemgeneration}.
Although we observe different verification pattern preferences by choosing polymath samples, we do not observe direct connection between the self-verification and the improvement in reasoning abilities. Besides, the polymath learning experiments are only conducted in open-ended format, while previous studies have demonstrated the benefits of incorporating diverse question-answer formats~\citep{akter2025nemotroncrossthinkscalingselflearningmath}, especially for benchmarks that are in multiple-choice formats. Moreover, our study mainly focuses polymath samples in math or employing math skills and does not extend to other domains where reliable rewards are accessible. 

\section{LIMR Score Basics}
\label{sec: LIMR score basics}
The LIMR score~\cite{li2025limrrlscaling} is computed by measuring the sample-wise training reward with the dataset-wise average. Specifically,

\begin{align*}
    s_{i} = 1 - \frac{\sum_{i=1}^{K}(r_{i}^{k} - \bar{r}^{k})^{2}}{\sum_{i=1}^{K}(1 - \bar{r}^{k})^{2}}, 
    \quad \bar{r}^{k} = \frac{1}{N}\sum_{i=1}^{N}r_{i}^{k}
\end{align*}
where $r_{i}^{k}$ is the reward of sample $i$ in the $k$-th epoch, and $\bar{r}^{k}$ is the average reward of training set in the $k$-th epoch.

\section{Configurations}
\label{sec: other config}

We employ a learning rate of 1e-6 during training, with $\epsilon$ being 0.2. The maximum generation length is set to 2048. The configuration to collect zero-shot sampling for base model is listed in Table~\ref{tab: zeroshot hyperparameter}. The prompt used is displayed in Table~\ref{tab: prompt-train}, and the prompt to synthesize polymath samples is shown in Table~\ref{tab:synthetic_prompt}. Around 500 candidate problems are synthesized on the candidate problem generation stage. The prompt employed for math skill identification is displayed in Table~\ref{tab: skill identification prompt}.  

\begin{table}[ht]
    \small
    \centering
    \begin{tabular}{c|l}
    \toprule
        \Thead{Hyperparameter} & \Thead{Value}\\
        \midrule
        temperature & 0.5 \\
        \midrule
        top k & 10 \\
        \midrule
        top p & 0.8 \\
        \bottomrule
    \end{tabular}
    \caption{Hyperparameters for computing 0-shot pass rate@k of the base model.}
    \label{tab: zeroshot hyperparameter}
\end{table}

\section{Polymath Learning in other RL approach}
\label{sec: polymath learning other RL}

Table~\ref{table: ppo results} demonstrates the results in employing PPO~\citep{schulman2017proximalpolicyoptimizationalgorithms} as the RL algorithm on Qwen2.5-7b-base with our \syntheticprime sample with other configurations being consistent. Our results show that polymath training with the \syntheticprime sample outperforms comprehensive learning with the MATH dataset.

\begin{table}[h!]
\centering
\resizebox{1.0\textwidth}{!}{
\begin{tabular}{lccccccccc}
\toprule
\textbf{Data} & \textbf{Math} & \textbf{Physics} & \textbf{Chemistry} & \textbf{Biology} & \textbf{Science} & \textbf{Engineering} & \textbf{Computer Science} & \textbf{Others} &  \textbf{Avg} \\
\midrule
0-shot & 20.4 & 4.4 & 4.4 & 5.1 & 0.0 & 3.7 & 3.3 & 9.6 & 6.4 \\
\midrule
MATH & 36.7 & 13.8 & 12.9 & 32.4 & 41.2 & 48.3 & 14.3 & 10.0 & 26.2 \\
\midrule
Prime & 37.7 & 17.9 & 14.2 & 35.3 & 48.5 & 57.6 & 17.1 & 13.9 & 30.3 \\
\bottomrule
\end{tabular}}
\caption{Comparison between polymath learning and comprehensive learning in PPO.}
\label{table: ppo results}
\end{table}

\begin{table*}[h!]
\begin{tcolorbox}[colback=blue!5!white,colframe=black,width=1.0\textwidth,title={Prompt for Training}]
\small
A conversation between User and Assistant. The user asks a question, and the Assistant solves it. The assistant first thinks about the reasoning process in the mind and then provides the user with the answer. User: You must put your answer inside \textbackslash \textbackslash boxed\{\} and Your final answer will be extracted automatically by the \textbackslash \textbackslash boxed\{\} tag. For multiple choice questions, the final answer in \textbackslash \textbackslash boxed\{\} should be the option letter (A, B, C, D, etc.).

[PROBLEM]

Assistant:
\end{tcolorbox}
\vspace{-0.3cm}
\caption{Training prompt, where [PROBLEM] is the placeholder for the problem.}
\label{tab: prompt-train}
\end{table*}

\vspace{-0.2cm}
\begin{table*}[h!]
\begin{tcolorbox}[colback=blue!5!white,colframe=black,width=1.0\textwidth,title={Prompt for Synthesizing Polymath Sample}]
\small
You are a professor proficient in physics, chemistry, and biology, tasked with creating a highly integrated problem for students that encompasses knowledge from all three disciplines. This problem should be a fill-in-the-blank question, and the final answer must be a precise integer (a positive integer between 1-1000). The difficulty of this question should be at the high school to university level. Furthermore, it should not involve any estimation, and complex calculations should be avoided as much as possible to ensure the robustness of the evaluation.
\end{tcolorbox}
\vspace{-0.3cm}
\caption{Prompt for synthesizing polymath sample.}
\label{tab:synthetic_prompt}
\end{table*}

\begin{table*}[ht!]
\begin{tcolorbox}[colback=blue!5!white,colframe=black,width=1.0\textwidth,title={Prompt for Skill Identification}]
\small
Here is a reasoning problem, and your job is to identify the concepts and skills in the scope of [CATEGORY] that are related to solve the problem.

Please separate the concepts or skills with ;, and if there is no skills or concepts identified, please answer with None. Please put your answer within $<$answer$><$/answer$>$.

For example: compute derivatives is the skill in precalculus.

Question:

[QUESTION]

\end{tcolorbox}
\caption{Prompt for skill identification. The [CATEGORY] and [QUESTION] are the placeholder for math category (e.g. algebra) and problem respectively.}
\label{tab: skill identification prompt}
\end{table*}

\section{Disentangling the Synthetic Prime Sample}
\label{sec: synthetic prime variants}

To better understand the effectiveness of the \syntheticprime sample that combines mathematical reasoning with multidisciplinary scientific context, we construct two variants that isolate these factors.

\begin{itemize}
    \item \textbf{Math-only} (Table~\ref{tab: polymath-shot-synthetic-math-only}): We convert the \syntheticprime sample into a purely mathematical problem by removing the multidisciplinary scientific context while preserving the underlying mathematical structure. The chemical and biology components are substituted with symbolic representations. The original units are retained as clues to infer the required reasoning formulas.
    \item \textbf{Science-only} (Table~\ref{tab: polymath-shot-synthetic-science-only}): We decompose the \syntheticprime sample into 3 milestone sub-problems, and provide the intermediate results as additional context to reduce the complexity of math reasoning while preserving the scientific context.
\end{itemize}

We compare the results of training with the two variants and the original \syntheticprime sample in Table~\ref{table: synthetic prime variants results}. Both variants achieve reasoning gains over training on the MATH dataset. Notably, the science-only variant, whose solution requires only simple arithmetic once intermediate values are identified, still leads to substantial improvements in multidiscipline reasoning. One possible explanation is that, despite reduced computational complexity, the problem retains reasoning components such as information selection, discipline-specific grounding, and structured dependencies, which may provide useful training signals other than mathematical computation. The math-only variant is able to achieve on-par performance with the \syntheticprime sample on several domains. These findings suggest that both the non-trivial math reasoning and structured subject-specific problem formulation contribute to the effectiveness of the \syntheticprime.

\begin{table*}[ht!]
\begin{tcolorbox}[colback=blue!5!white,colframe=black,width=1.0\textwidth,title={Polymath Sample (Synthetic Prime, Math-only)}]
\small
\examplequestion{[Question]}
You are analyzing a system defined by a sequence of 11 symbols: $S$ = 'G C G C G C G C A T A'. \\

Each symbol contributes a specific 'cost unit' to the total system load:\\
- Symbols G or C contribute 3 units.\\
- Symbols A or T contribute 2 units.\\
- Let $T$ be the sum of all cost units in sequence $S$.\\

Here are the relevant constants and their dimensions to this problem\\
- $C_1 = 6.626 \times 10^{-34} \text{J · s}$\\
- $C_2 = 3.00 \times 10^{8} \text{m · s}$\\
- $C_3 = 400 \times 10^{-9} \text{m}$\\
- $C_4 = 20 \times 10^{3} \text{J · mol}^{-1}$\\
- $C_5 = 6.022 \times 10^{23} \text{mol}^{-1}$\\

You must compute two energy expressions $E_{packet}$ and $E_{unit}$, both with the dimension Joules (J). Note that\\
- $E_{\text{packet}}$ must be formed using **only** $C_{1}$, $C_{2}$, $C_{3}$.\\
- $E_{\text{unit}}$ must be formed using **only** $C_{4}$, $C_{5}$.\\
- Assume all the dimensionless numerical constant is **1**.\\

**Fill in the blank:** Please find the minimum integer $n$ such that $n \times E_{\text{packet}} \geq T \times E_{\text{unit}}$.  (Answer with a single positive integer.)

\examplequestion{[Answer]}
\( \boxed{2} \)
\end{tcolorbox}
\vspace{-0.2cm}
\caption{The \syntheticprime sample with math-only context.}
\label{tab: polymath-shot-synthetic-math-only}
\end{table*}

\begin{table}[h]
\centering
\resizebox{1.0\textwidth}{!}{
\begin{tabular}{lccccccccc}
\toprule
\textbf{Data} & \textbf{Math} & \textbf{Physics} & \textbf{Chemistry} & \textbf{Biology} & \textbf{Science} & \textbf{Engineering} & \textbf{Computer Science} & \textbf{Others} &  \textbf{Avg} \\
\midrule
MATH & 37.2 & 12.8 & 10.0 & 31.4 & 6.5 & 8.6 & 25.8 & 23.4 & 19.5 \\
\midrule
Prime (Science-only) & 37.0 & 14.8 & 12.4 & 50.0 & 11.3 & 16.5 & 41.2 & 29.6 & 26.6 \\
\midrule
Prime (Math-only) & 36.4 & \underline{20.3} & \underline{16.1} & 52.5 & 12.5 & 19.0 & 45.4 & \underline{32.9} & 29.4 \\
\midrule
Prime & \textbf{38.3} & \underline{20.6} & \underline{15.7} & \textbf{54.2} & \textbf{15.6} & \textbf{20.8} & \textbf{48.5} & \underline{32.4} & \textbf{30.8} \\
\bottomrule
\end{tabular}}
\caption{Comparison between polymath learning with the \syntheticprime sample and the variants that decouple the math or scientific context. The best performance is \textbf{bolded} and compariable performance is \underline{underlined}.}
\label{table: synthetic prime variants results}
\end{table}

\begin{table*}[ht!]
\begin{tcolorbox}[colback=blue!5!white,colframe=black,width=1.0\textwidth,title={Polymath Sample (Synthetic Prime, Science-only)}]
\small
\examplequestion{[Question]}
A double‑stranded DNA fragment of exactly 11 base pairs has the upper strand sequence\\
5 'G C G C G C G C A T A 3'.\\

Each adenine–thymine (A·T) base pair is held together by **2** hydrogen bonds, and each guanine–cytosine (G·C) base pair by **3** hydrogen bonds. \textcolor{red}{Therefore, there are 30 hydrogen bonds in total.}

The DNA molecule is irradiated with monochromatic light of wavelength $\lambda = 400 \text{nm}$, \textcolor{red}{and the energy of each photon is $4.97 \times 10^{-19}$ J}.
Assume that **100\%** of every photon’s energy is used exclusively to break hydrogen bonds between the two strands. \textcolor{red}{And it requires $3.32 \times 10^{-20}$ J to break one bond.}

Use the exact data below (treat every value as exact):

* Enthalpy of one hydrogen bond $\Delta H = 20 \text{kJ · mol}^{-1}$\\  
* Planck constant $h = 6.626 \times 10^{-34} \text{J · s}$ \\
* Speed of light $c = 3.00 \times 10^{8} \text{m· s}^{-1}$ \\ 
* Avogadro constant $N_{a} = 6.022 \times 10^{23} \text{mol}^{-1}$ \\

**Fill in the blank:**  
What is the minimum number of 400 nm photons required to supply exactly enough energy to dissociate *all* hydrogen bonds in **one** molecule of this DNA fragment?  
(Answer with a single positive integer.)\\

\examplequestion{[Answer]}
\( \boxed{2} \)
\end{tcolorbox}
\vspace{-0.2cm}
\caption{The \syntheticprime sample with science-only context, the additional intermediate context is \textcolor{red}{highlighted}.}
\label{tab: polymath-shot-synthetic-science-only}
\end{table*}

\section{Sample Preference with LIMR Scores}
\label{sec: sample preference with LIMR score}
\begin{figure}[h!]
    \centering
    \includegraphics[width=0.4\linewidth]{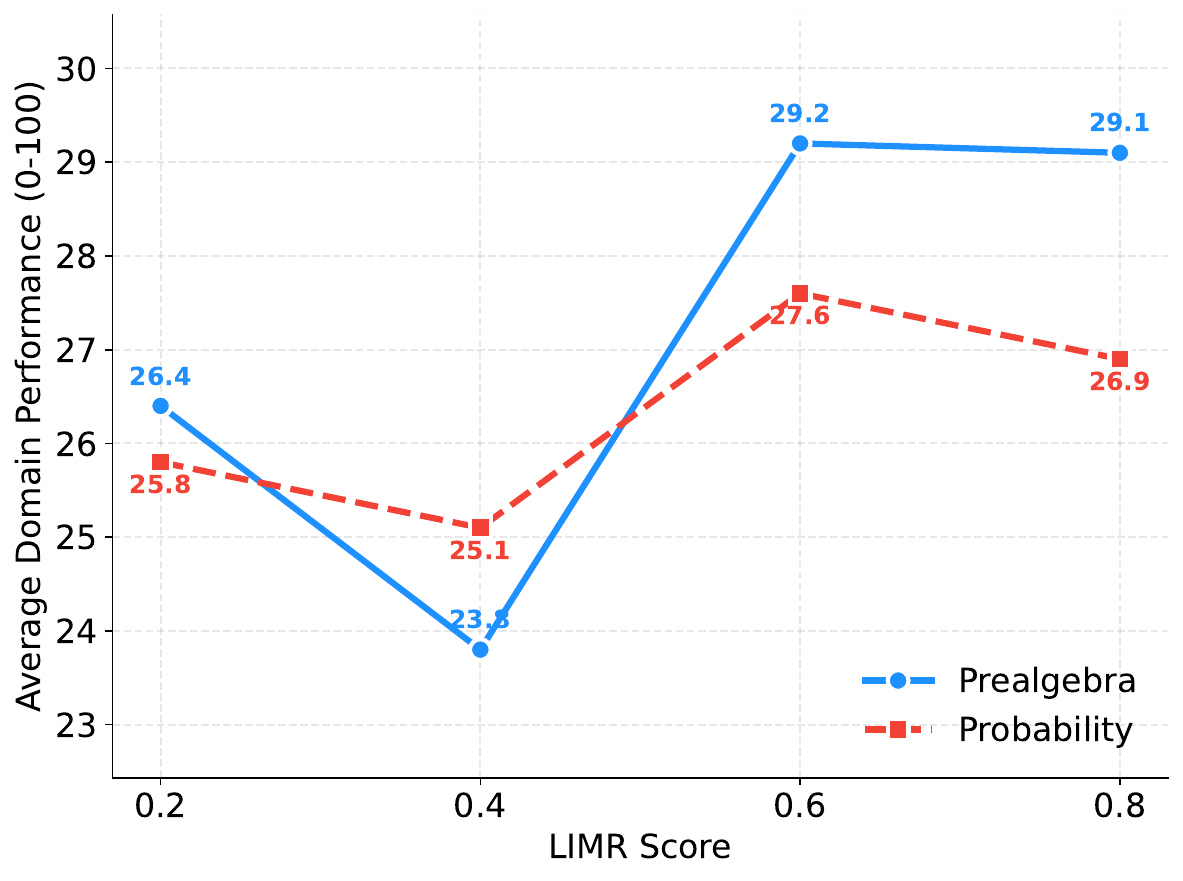}
    \caption{Average domain performance over natural samples with different LIMR scores. The performance is reported the same way as in Table~\ref{table: main results by subject domains}. The samples with LIMR score being 0.6 perform the strongest.}
    \label{fig: limr score selection}
\end{figure}

We include the results from selecting different LIMR scores from two math categories, prealgebra and probability. The results in Figure~\ref{fig: limr score selection} show that the samples with LIMR score being 0.6 deliver the best performance.

\section{Results by Datasets}
\label{sec: results by datasets}
Table~\ref{table: main results by benchmark} includes results by datasets on polymath learning and comprehensive learning, with the synthetic sample still performing the strongest.

\begin{table}[h]
\centering
\caption{Results on different reasoning benchmarks, where \textit{OE} refers to benchmarks of open-ended problems: MATH500, AIME2024, AIME2025, Minerva and Scibench, while \textit{MCQ} refers to benchmarks of multiplechoice problems. The best performance is bolded and the best polymath learning performance is underlined if not optimal.}
\label{table: main results by benchmark}
\resizebox{1.0\textwidth}{!}{
\begin{tabular}{lccccccccccc}
\toprule
\textbf{Polymath Subject} & \textbf{MATH500} & \textbf{AIME2024} & \textbf{AIME2025} & \textbf{Minerva} & \textbf{GPQA-Diamond} & \textbf{SuperGPQA} & \textbf{MMLU-Pro} & \textbf{SciBench} & \textbf{AVG-OE} & \textbf{AVG-MCQ} & \textbf{AVG-All} \\
\rowcolor{lightblue}
\multicolumn{12}{c}{\textbf{N=64 Sampling (0 shot)}}\\
- & 54.8 & 9.0 & 7.1 & 13.4 & 13.1 & 15.7 & 4.7 & 9.8 & 23.6 & 11.3 & 15.9\\
\rowcolor{lightblue}
\multicolumn{12}{c}{\textbf{In-context Learning (1 shot)}}\\
\rowcolor{SampleSeparator}
\multicolumn{12}{c}{\textbf{Natural Sample}}\\
Geometry & 60.0 & 8.2 & 4.7 & 15.4 & 9.6 & 4.5 & 20.5 & 6.8 & 19.0 & 11.5 & 16.2 \\
Prealgebra & 55.0 & 9.2 & 4.5 & 10.7 & 16.2 & 9.2 & 28.8 & 6.4 & 17.2 & 18.1 & 17.5 \\
Algebra & 48.0 & 8.2 & 3.1 & 15.8 & 14.6 & 10.7 & 25.6 & 6.7 & 16.4 & 17.0 & 16.6 \\
Intermediate Algebra & 59.6 & 5.1 & 4.5 & 12.1 & 14.1 & 7.3 & 20.5 & 5.7 & 17.4 & 14.0 & 16.1 \\
Number Theory & 52.8 & 8.5 & 3.9 & 11.8 & 16.7 & 6.3 & 23.4 & 5.9 & 16.6 & 15.5 & 16.2 \\
Precalculus & 51.8 & 6.7 & 3.9 & 15.8 & 13.1 & 4.9 & 19.0 & 5.2 & 16.7 & 12.3 & 15.0 \\
Probability & 54.2 & 7.3 & 4.0 & 13.6 & 11.1 & 6.3 & 19.7 & 5.8 & 17.0 & 12.4 & 15.2 \\
\rowcolor{SampleSeparator}
\multicolumn{12}{c}{\textbf{Synthetic Sample}}\\
Prime & 44.2 & 4.8 & 2.4 & 15.1 & 5.6 & 2.8 & 10.6 & 3.8 & 14.1 & 6.3 & 11.2 \\
\rowcolor{lightblue}
\multicolumn{12}{c}{\textbf{Comprehensive Learning ($> \text{1k shots}$)}}\\
\rowcolor{SampleSeparator}
\multicolumn{12}{c}{\textbf{Natural Sample}}\\
MATH (8k) & 73.6 & 13.0 & 7.9 & \textbf{30.9} & 11.7 & 10.3 & 22.5 & \textbf{23.1} & 29.7 & 14.8 & 24.1 \\
LIMR (1k) & \textbf{74.8} & 12.6 & 8.9 & 30.1 & 13.2 & 15.8 & 31.5 & 22.7 & 29.8 & 20.2 & 26.2 \\
\rowcolor{lightblue}
\multicolumn{12}{c}{\textbf{Polymath Learning (1 shot)}}\\
\rowcolor{SampleSeparator}
\multicolumn{12}{c}{\textbf{Natural Sample}}\\
Geometry & 26.6 & 0.0 & 0.0 & 19.9 & \textbf{23.9} & 18.5 & 33.1 & 7.9 & 10.9 & 25.2 & 16.2 \\
Prealgebra & 71.2 & 13.3 & 13.3 & 30.9 & 18.3 & 19.4 & 35.0 & 21.4 & 30.0 & 24.2 & \textbf{27.9} \\
Algebra & 72.0 & 6.7 & 0.0 & \textbf{30.9} & 16.2 & 17.3 & 34.9 & \underline{22.8} & 26.5 & 22.8 & 25.1 \\
Intermediate Algebra & 71.2 & 13.3 & 0.0 & 28.7 & 20.3 & 18.9 & 34.5 & 22.0 & 27.0 & 24.6 & 26.1 \\
Number Theory & 69.6 & \textbf{16.7} & 10.0 & \textbf{30.9} & 17.8 & 18.2 & 35.0 & 22.3 & 29.9 & 23.7 & 27.6 \\
Precalculus & 71.6 & 10.0 & 10.0 & 30.5 & 18.8 & 20.9 & 34.1 & 22.4 & 28.9 & 24.6 & 27.3 \\
Probability & 71.6 & 13.3 & \textbf{16.7} & 29.8 & 14.2 & 18.9 & 34.9 & 22.7 & \textbf{30.8} & 22.7 & \textbf{27.8} \\
\rowcolor{SampleSeparator}
\multicolumn{12}{c}{\textbf{Synthetic Sample}}\\
Geometry & 71.4 & 10.2 & 6.7 & 27.2 & 15.7 & 16.9 & 30.7 & 21.4 & 27.4 & 21.1 & 25.0\\
Algebra & 71.6 & 10.2 & 6.7 & \textbf{30.9} & 20.3 & 19.3 & 33.6 & 21.8 & 28.2 & 24.4 & 26.8 \\
Number Theory & \underline{73.8} & 11.7 & 7.1 & 29.8 & 14.2 & 19.3 & 34.6 & \textbf{23.1} & 29.1 & 22.7 & 26.7 \\
Precalculus & 71.8 & 11.4 & 7.7 & 29.4 & 19.8 & \textbf{21.5} & 35.8 & \underline{22.8} & 28.6 & 25.7 & 27.5 \\
Probability & 71.8 & 11.6 & 7.2 & 28.3 & 16.8 & 17.5 & 36.4 & 22.1 & 28.2 & 23.6 & 26.5\\
\midrule
Prime & 71.4 & 10.1 & 7.2 & \textbf{30.9} & 21.3 & 20.5 & \textbf{38.4} & 22.3 & 28.4 & \textbf{26.7} & \textbf{27.8} \\
\bottomrule
\end{tabular}}
\end{table}

\section{Full Subject List}
\label{sec: full eval subject list}
The full list of reasoning subjects being evaluated is displayed in Table~\ref{tab: evaluation benchmarks}.
\begin{table*}[h]
    \centering
    \resizebox{0.8\textwidth}{!}{
    \begin{tabular}{|>{\centering\arraybackslash}p{3cm}|>{\raggedright\arraybackslash}p{3cm}|c|c|c}
    \toprule
    \Thead{Subject Domain} & \Thead{Subject} & \Thead{Source} & \Thead{\# Samples}\\
    \midrule
    Math & AIME & AIME2024, AIME2025 & 60 \\
    & MATH500 & MATH & 500 \\
    & Minerva & MinervaMath & 272 \\
    & math & Scibench, MMLU-Pro & 299 \\
    \midrule
    Physics & physics & GPQA-Diamond, Scibench, MMLU-Pro & 413\\
    \midrule
    Chemistry & chemistry & GPQA-Diamond, Scibench, MMLU-Pro & 459\\
    \midrule
    Biology & biology & GPQA-Diamond, Scibench, MMLU-Pro & 118\\
    \midrule
    Science & science & SuperGPQA & 557 \\
    \midrule
    Engineering & engineering & SuperGPQA & 447 \\
    \midrule
    Computer Science & computer science & MMLU-Pro & 100\\
    \midrule
    Others & military science & SuperGPQA & 12 \\
    & business & MMLU-Pro & 100 \\
    & philosophy & MMLU-Pro, SuperGPQA & 120 \\
    & economics & MMLU-Pro, SuperGPQA & 149 \\
    & management & SuperGPQA & 28 \\
    & health & MMLU-Pro & 100 \\
    & psychology & MMLU-Pro & 100 \\
    & medicine & SuperGPQA & 155 \\
    & education & SuperGPQA & 27 \\
    & agronomy & SuperGPQA & 27 \\
    & literature and arts & SuperGPQA & 93 \\
    & law & MMLU-Pro, SuperGPQA & 137 \\
    & history & MMLU-Pro, SuperGPQA & 138 \\
    & sociology & SuperGPQA & 8 \\
    & other & MMLU-Pro & 100 \\
    \bottomrule
    \end{tabular}
}
    \caption{Evaluation reasoning benchmarks with subjects included.}
    \label{tab: evaluation benchmarks}
\end{table*}

\section{Robustness of Experiments}
\label{sec: result robustness}
We include the results of comprehensive learning in MATH train set and polymath learning with the \syntheticprime sample in 3 independent runs on Qwen2.5-7b-base. The results in Table~\ref{table: robustness subject domains} shows that the comprehensive learning on 8k MATH samples demonstrate stronger reasoning in math benchmarks, but polymath learning with the \syntheticprime sample outperforms comprehensive learning on the MATH training set in most other benchmarks as well as the overall performance.

\begin{table}[h]
\centering
\resizebox{1.0\textwidth}{!}{
\begin{tabular}{lccccccccccc}
\toprule
\textbf{Polymath Subject} & \textbf{MATH500} & \textbf{AIME2024} & \textbf{AIME2025} & \textbf{Minerva} & \textbf{GPQA-Diamond} & \textbf{SuperGPQA} & \textbf{MMLU-Pro} & \textbf{SciBench} & \textbf{AVG-OE} & \textbf{AVG-MCQ} & \textbf{AVG-All} \\
\rowcolor{lightblue}
\multicolumn{12}{c}{\textbf{Comprehensive Learning ($> \text{1k shots}$)}}\\
MATH (8k) & \textbf{73.0}$\pm$0.59 & \textbf{15.6}$\pm$4.16 & 6.7$\pm$0.0 & 29.5$\pm$1.24 & 11.9$\pm$0.24 & 11.6$\pm$1.75 & 25.0$\pm$2.94 & \textbf{23.5}$\pm$0.37 & \underline{29.7}$\pm$0.73 & 16.2$\pm$1.53 & 24.6$\pm$0.72 \\
\rowcolor{lightblue}
\multicolumn{12}{c}{\textbf{Polymath Learning (1 shot)}}\\
Prime & 71.7$\pm$0.34 & 12.2$\pm$1.56 & \textbf{10.0}$\pm$4.71 & \textbf{31.0}$\pm$1.07 & \textbf{20.3}$\pm$0.71 & \textbf{20.8}$\pm$0.31 & \textbf{38.1}$\pm$0.69 & 21.9$\pm$0.33 & \underline{29.4}$\pm$1.03 & \textbf{26.4}$\pm$0.29 & \textbf{28.2}$\pm$0.62\\
\bottomrule
\end{tabular}}
\caption{The results of comprehensive learning on MATH and polymath learning on the \syntheticprime sample with 3 independent runs in Qwen2.5-7b-base. The best performance is \textbf{bold} as the on par performance is \underline{underlined}. Polymath learning on the \syntheticprime sample outperforms comprehensive learning with MATH on most benchmarks as well as the overall performance.}
\label{table: robustness subject domains}
\end{table}

\section{Self-verification by Subject Domains}
\label{sec: self-verification by sbuject domain}
\begin{figure}[h]
    \centering
    \includegraphics[width=0.99\linewidth]{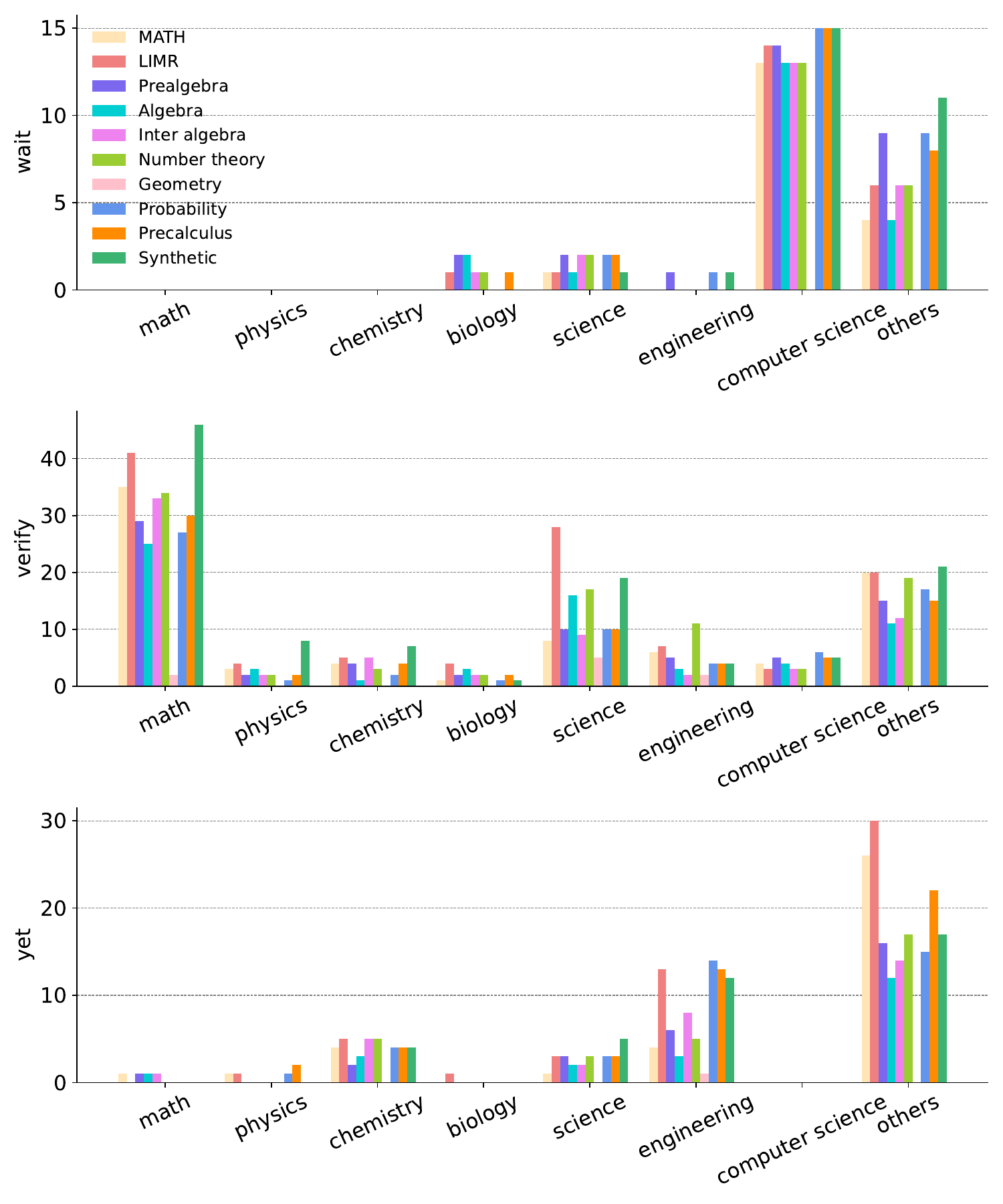}
    \caption{The verification patterns identified for `wait', `verify' and `yet' in different subject groups. The `wait' rates in computer science problems are highly attributed from terms in the question stems.}
    \label{fig:self verification subject group 1}
\end{figure}

\begin{figure}[h]
    \centering
    \includegraphics[width=0.99\linewidth]{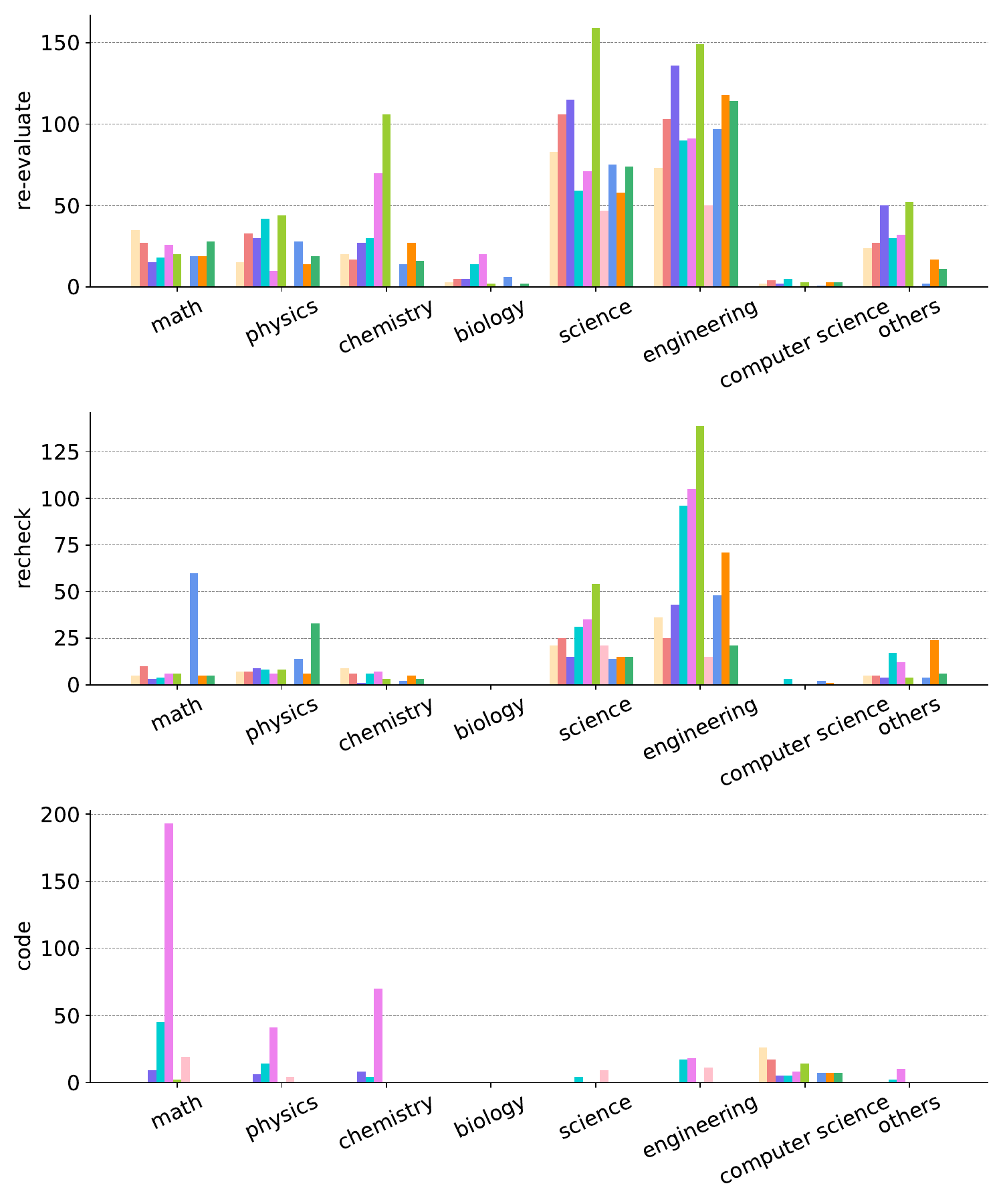}
    \caption{The verification patterns identified for `re-evaluate', `recheck' and `code' in different subject groups.}
    \label{fig:self verification subject group 2}
\end{figure}

We list the self-verification statistics by different sbuject domains in Figure~\ref{fig:self verification subject group 1} and Figure~\ref{fig:self verification subject group 2}. Specifically, we found that `verify' is more preferred in math problems while `re-evaluate' appears more frequently in science and engineering. Besides, polymath learning with the `intermediate algebra' sample elicits the most coding verifications among all the natural and synthetic samples.

\clearpage

\section{Polymath Learning with Other 1-shot Sample}
\label{sec: pi_1}
\begin{figure}[h]
    \centering
    \includegraphics[width=0.6\linewidth]{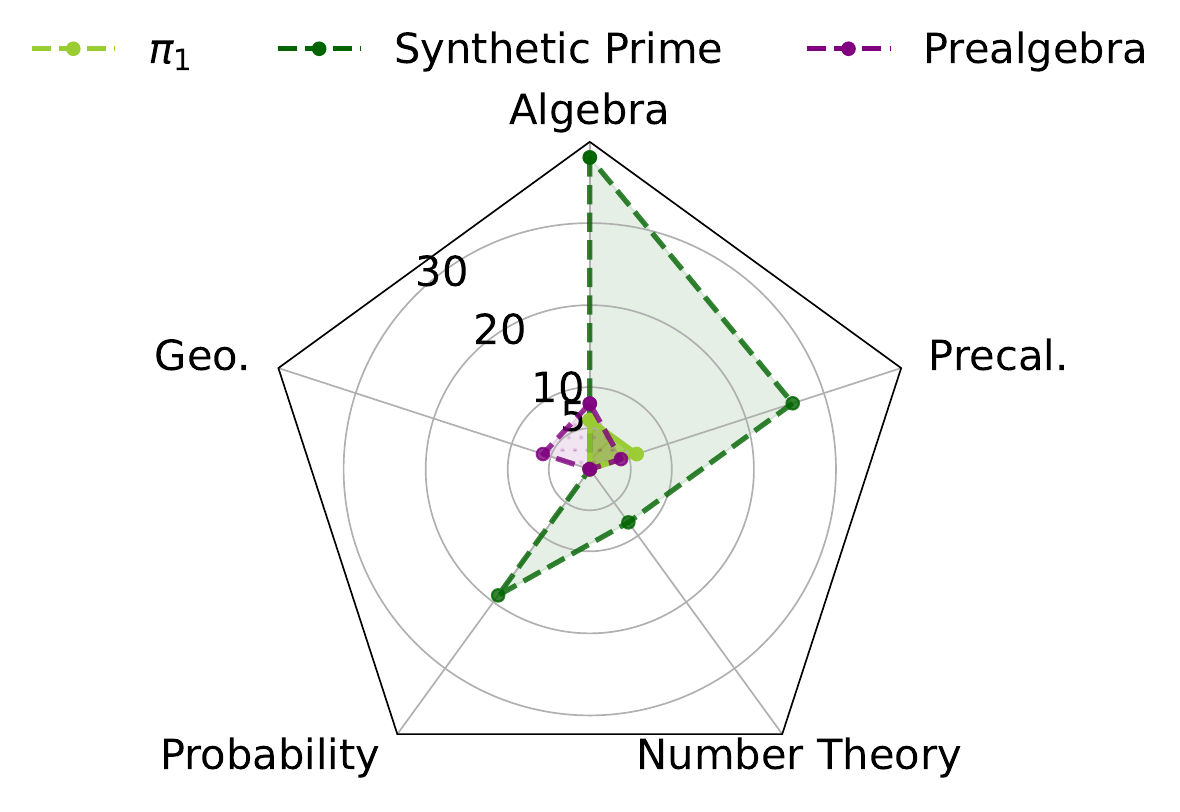}
    \caption{The skill spectrum between the $\pi_{1}$ sample, the \syntheticprime sample, and the strongest natural polymath sample in prealgebra. The strongest natural polymath and synthetic samples demonstrate richer and more comprehensive skill coverage than the $\pi_{1}$ sample.}
    \label{fig: skill radar pi 1}
\end{figure}

$\pi_{1}$ (see Table~\ref{tab: pi_1-sample}) is employed in previous success of reinforcement learning with one sample~\citep{wang2025reinforcement, wang2025unleashing}. It is selected from DeepScaleR~\citep{deepscaler2025}, a curated dataset of challenging mathematical competition problems like AIME and Omni-math~\citep{gao2025omnimath} other than MATH. Results in Table~\ref{table: domain results pi_1} demonstrate the effectiveness of \syntheticprime sample over both $\pi_{1}$ and comprehensive learning with 8k MATH samples in Qwen2.5-base in both 7b and 14b sizes. The skill abundance comparison with the strong synthetic and natural polymath sample (\syntheticprime sample and prealgebra) in Figure~\ref{fig: skill radar pi 1} also demonstrates more complex skill combinations than $\pi_{1}$ to solve.

\begin{table}[h]
\centering
\caption{The results between comprehensive learning on 8k MATH samples and polymath learning on the \syntheticprime sample and $\pi_{1}$ in Qwen2.5-7b-base and Qwen2.5-14b-base. The \syntheticprime sample consistently outperforms the other two data choices across models.}
\label{table: domain results pi_1}
\resizebox{1.0\textwidth}{!}{
\begin{tabular}{lccccccccc}
\toprule
\textbf{Data} & \textbf{Math} & \textbf{Physics} & \textbf{Chemistry} & \textbf{Biology} & \textbf{Science} & \textbf{Engineering} & \textbf{Computer Science} & \textbf{Others} &  \textbf{Avg} \\
\rowcolor{SampleSeparator}
\multicolumn{10}{c}{\textbf{Qwen2.5-7b-base}}\\
\rowcolor{lightblue}
\multicolumn{10}{c}{\textbf{N=64 Sampling (0-shot)}}\\
- & 20.4 & 4.4 & 4.4 & 5.1 & 0.0 & 3.7 & 3.3 & 9.6 & 6.4 \\
\rowcolor{lightblue}
\multicolumn{10}{c}{\textbf{Comprehensive Learning ($>$ 1k shots)}}\\
MATH (8k) & 37.2 & 12.8 & 10.0 & 31.4 & 6.5 & 8.6 & 25.8 & 23.4 & 19.5 \\
\rowcolor{lightblue}
\multicolumn{10}{c}{\textbf{Polymath Learning (1-shot)}}\\
$\pi_{1}$ (DeepScaleR) & 35.5 & 14.3 & 11.3 & 28.4 & \textbf{35.1} & \textbf{44.1} & 13.8 & 10.4 & 24.1 \\
Prime & \textbf{38.3} & \textbf{20.6} & \textbf{15.7} & \textbf{54.2} & 15.6 & 20.8 & \textbf{48.5} & \textbf{32.4} & \textbf{30.8} \\
\rowcolor{SampleSeparator}
\multicolumn{10}{c}{\textbf{Qwen2.5-14b-base}}\\
\rowcolor{lightblue}
\multicolumn{10}{c}{\textbf{N=64 Sampling (0-shot)}}\\
- & 37.7 & 26.2 & 22.2 & 28.1 & 41.2 & 39.0 & 20.8 & 14.3 & 28.7 \\
\rowcolor{lightblue}
\multicolumn{10}{c}{\textbf{Comprehensive Learning ($>$ 1k shots)}}\\
MATH (8k) & 42.7 & 26.4 & 20.5 & \textbf{44.7} & 49.5 & \textbf{64.4} & 22.3 & 15.6 & 35.8 \\
\rowcolor{lightblue}
\multicolumn{10}{c}{\textbf{Polymath Learning (1-shot)}}\\
$\pi_{1}$ (DeepScaleR) & 40.4 & 27.6 & 20.0 & 39.4 & 51.5 & 57.6 & 22.1 & 17.1 & 34.5 \\
Prime & \textbf{44.0} & \textbf{32.7} & \textbf{22.7} & 42.3 & \textbf{56.7} & 58.5 & \textbf{31.0} & \textbf{20.6} & \textbf{38.6}\\
\bottomrule
\end{tabular}}
\end{table}

\section{Performance on MMLU-Pro and SuperGPQA Full Set}
\label{sec: mmlu-pro and superGPQA full}
Table~\ref{table: MMLU-Pro and SuperGPQA full} reports the results on full MMLU-Pro and SuperGPQA for comprehensive learning and strong polymath samples trained with Qwen2.5-7B-Base under the same configuration described in Section~\ref{sec: experimental setup}. Polymath learning on the \syntheticprime sample achieves substantially higher performance than both 0-shot learning and comprehensive learning using thousands of samples.

\begin{table}[ht]
\centering
\caption{Performance of different comprehensive learning and polymath learning samples on the full set of MMLU-Pro and SuperGPQA, the \syntheticprime sample performs best (\textbf{bolded}).}
\label{table: MMLU-Pro and SuperGPQA full}
\resizebox{0.5\textwidth}{!}{
\begin{tabular}{lcc}
\toprule
\textbf{Data} & \textbf{MMLU-Pro$^{\text{full}}$} & \textbf{SuperGPQA$^{\text{full}}$} \\
\midrule
0-shot & 30.3 & 16.8 \\
MATH (8k) & 31.7 & 16.6 \\
LIMR (1k) & 33.0 & 17.2 \\
$\pi_{1}$ & 29.7 & 16.7 \\
Prealgebra & 33.4 & 19.2 \\
Prime & \textbf{37.6} & \textbf{21.7} \\
\bottomrule
\end{tabular}}
\end{table}

\section{Training Dynamics of Polymath Learning}
\label{sec: training dynamics}
\begin{figure}[h]
    \centering
    \includegraphics[width=0.99\linewidth]{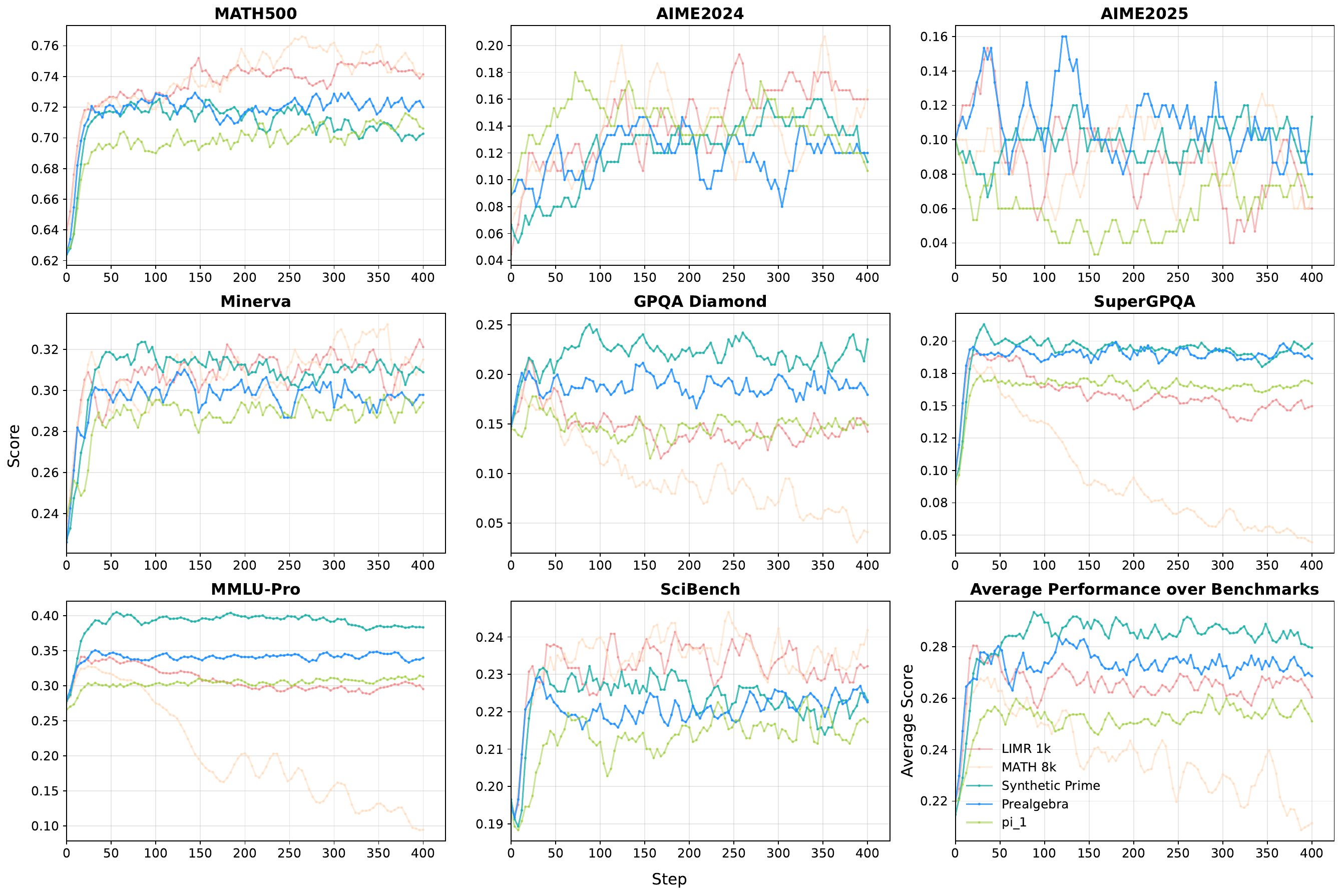}
    \caption{The evaluation results of benchmarks between comprehensive learning (MATH and LIMR) and different polymath learning samples (\syntheticprime sample, natural prealgebra sample, $\pi_{1}$) trained in Qwen2.5-7b-base. The results are collected in greedy decoding and rolling smoothing average with window of 5 is applied to AIME2024, AIME2025 and 3 for other benchmarks for demonstration purpose.}
    \label{fig: training convergence}
\end{figure}

Figure~\ref{fig: training convergence} illustrates the training dynamics of comprehensive learning and polymath learning across strong natural and synthetic samples. We specifically prolong the training to better observe convergence. We observe that comprehensive learning, on either the 8k MATH training set or the LIMR subset, yields progressive improvement on MATH500, but exhibits pronounced overfitting on multidisciplinary benchmarks such as GPQA Diamond, SuperGPQA, and MMLU-Pro. And training with the MATH set exacerbates this effect. Polymath learning, on the other hand, demonstrates substantially greater robustness especially on multidisciplinary reasoning benchmarks even though demonstrate inferior performance on MATH500 compared to comprehensive learning. Moreover, both the \syntheticprime sample and natural polymath sample in prealgebra deliver stronger multidisciplinary reasoning performance than the $\pi_{1}$ sample.

\section{Polymath Learning on Additional Models}
\label{sec: additional model polymath learning}

Table~\ref{table: additional results by benchmark} includes comparison between comprehensive learning and polymath learning in additional model choices. Specifically, we select Qwen2.5-14b-base, Qwen3-8b-base~\citep{yang2025qwen3technicalreport} and OctoThinker-8b-long-base~\citep{wang2025octothinkermidtrainingincentivizesreinforcement}, which strengthens reasoning ability of Llama3.2~\citep{llama32} through mid-training in long-form reasoning data. The results show that the effectiveness of polymath learning on the \syntheticprime extends to most models. Although it does not surpass comprehensive learning in OctoThinker-8b-long-base, it nonetheless yields improvements in multidiscipline reasoning benchmarks (GPQA-Diamond, SuperGPQA). This trend echoes observations in~\citet{dohmatob2025sometimestheorydatacuration} regarding the relationship between data curation effectiveness and the ability of the underlying model.

\begin{table}[ht]
\centering
\resizebox{1.0\textwidth}{!}{
\begin{tabular}{lccccccccccc}
\toprule
\textbf{Data} & \textbf{MATH500} & \textbf{AIME2024} & \textbf{AIME2025} & \textbf{Minerva} & \textbf{GPQA-Diamond} & \textbf{SuperGPQA} & \textbf{MMLU-Pro} & \textbf{SciBench} & \textbf{AVG-OE} & \textbf{AVG-MCQ} & \textbf{AVG-All} \\
\rowcolor{SampleSeparator}
\multicolumn{12}{c}{\textbf{Qwen2.5-Math-7b (2k context)}}\\
0-shot & 68.6 & 16.7 & 10.0 & 27.2 & 11.7 & 8.0 & 11.4 & \underline{22.1} & 28.9 & 10.4 & 22.0 \\
MATH (8k) & \textbf{76.0} & \textbf{26.7} & 6.7 & \underline{29.0} & 11.7 & 9.0 & \underline{18.8} & 20.2 & \textbf{31.7} & 13.2 & \underline{24.8} \\
Prime  & 72.0 & 13.3 & \textbf{13.3} & \underline{29.8} & \textbf{14.2} & \textbf{11.5} & \underline{18.2} & \underline{22.1} & 30.1 & \textbf{14.6} & \underline{24.3} \\
\rowcolor{SampleSeparator}
\multicolumn{12}{c}{\textbf{Qwen2.5-14b-base (2k context)}}\\
0-shot & 68.6  & 16.7  & 3.3 & 26.8 & 29.9 & 16.8 & 42.2 & 19.7 & 27.0 & 29.6 & 28.0\\
MATH (8k) & \textbf{77.6} & \textbf{20.0}  & 6.7 & 34.2 & 28.4 & 23.4 & 46.9 & \textbf{27.2} & \textbf{33.1} & 32.9 &  33.1 \\
Prime & 76.0 & 16.7 & \textbf{10.0} & \textbf{35.3} & \textbf{37.1} & \textbf{26.1} & \textbf{53.3}  & 23.6 & 32.3 & \textbf{38.8} & \textbf{34.8} \\
\rowcolor{SampleSeparator}
\multicolumn{12}{c}{\textbf{Qwen3-8b-base (8k context)}}\\
0-shot & 58.8 & 10.0 & 6.7 & 28.3 & 23.4 & 20.9 & 35.9 & 16.9 & 24.1 & 26.7 & 25.1 \\
MATH (8k) & \textbf{79.8} & \underline{16.7} & 13.3 & 34.2 & 16.2 & 10.1 & 30.0 & \textbf{29.2} & \textbf{34.6} & 18.8 & 28.7 \\
Prime & 73.2 & \underline{16.7} & \textbf{16.7} & \textbf{35.3} & \textbf{35.0} & \textbf{25.0} & \textbf{51.9} & 25.4 & 33.5 & \textbf{37.3} & \textbf{34.9} \\
\rowcolor{SampleSeparator}
\multicolumn{12}{c}{\textbf{OctoThinker-8b-long-base (8k context)}}\\
0-shot & 8.6 & 3.3 & 0.0 & 9.6 & 0.0 & 0.1 & 0.4 & 2.0 & 4.7 & 0.2 & 3.0 \\
MATH (8k) & \textbf{73.0} & \textbf{16.7} & \textbf{13.3} & \textbf{22.4} & 17.8 & 16.3 & \textbf{41.5} & \textbf{22.0} & \textbf{29.5} & 25.2 & \textbf{27.9} \\
Prime & 14.0 & 0.0 & 0.0 & 11.8 & \textbf{28.4} & \textbf{17.1} & 33.0 & 5.8 & 6.3 & \textbf{26.2} & 13.8 \\
\bottomrule
\end{tabular}}
\caption{Performance of comprehensive learning on 8k MATH samples and the \syntheticprime sample on reasoning benchmarks with additional model choices. The best performance is \textbf{bolded} and the on-par performance is \underline{underlined}. The \syntheticprime sample outperforms comprehensive learning when trained with strong model like Qwen2.5-14b-base and in some non-math benchmarks when trained with OctoThinker-8b-long-base.}
\label{table: additional results by benchmark}
\end{table}

\section{Reasoning Breakdown by Subject}
\label{sec: reasoning breakdown by subject}
\begin{figure}[h]
    \centering
    \includegraphics[width=0.9\linewidth]{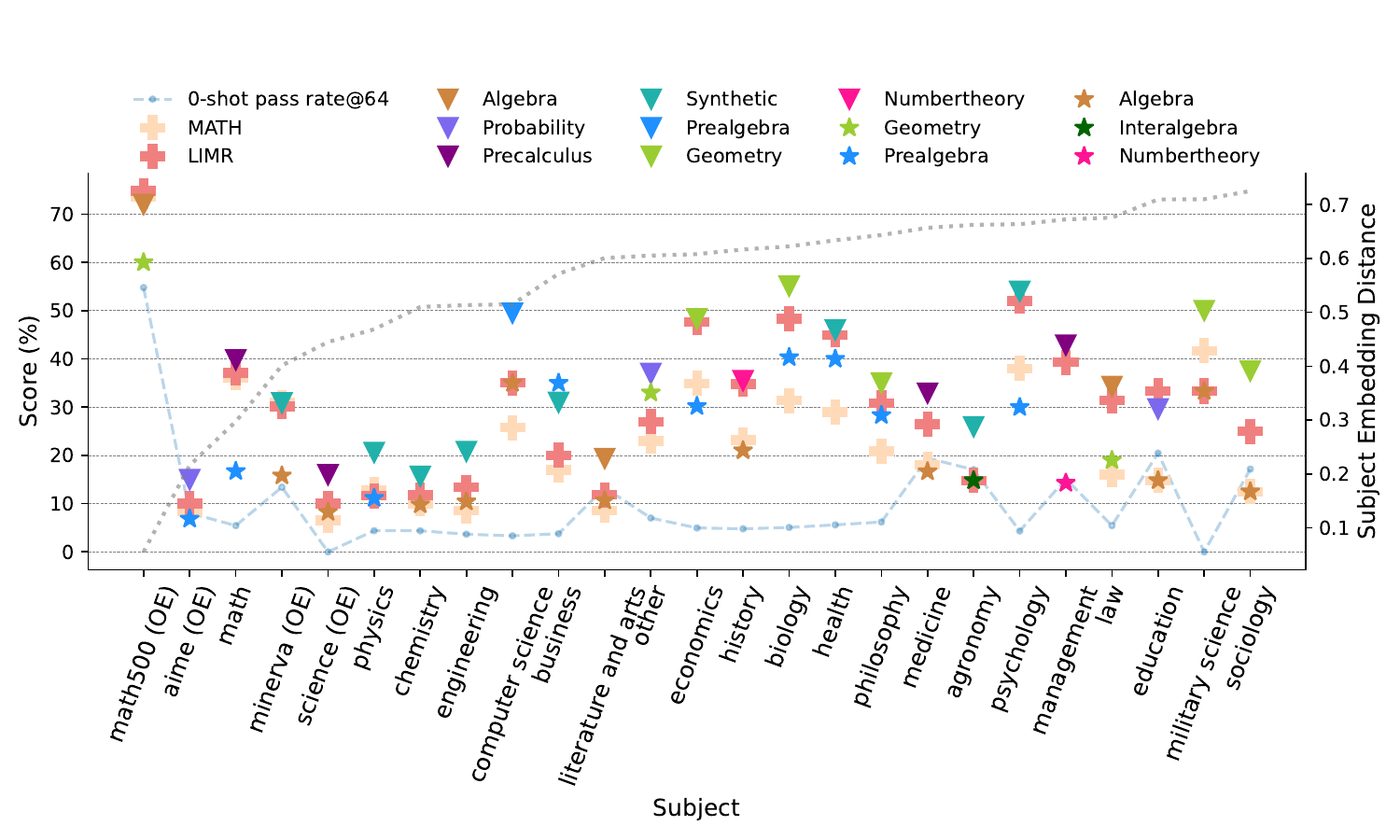}
    \caption{The subject-level performance of different learning strategies. \textit{OE} stands for subjects with open-ended problems. The subjects are sorted by subject embedding distance to MATH500 (the grey dotted line), from low to high. The blue line represents pass ratio from 64 independent attempts of the base model. The stars and triangles represent best performance of in-context learning and polymath learning. Note that we only display the best polymath learning and in-context polymath learning results for demonstration, and \textit{Synthetic} represents the \syntheticprime sample.}
    \label{fig:polymath learning perf by subject embedding - diff sample}
\end{figure}

Figure~\ref{fig:polymath learning perf by subject embedding - diff sample} illustrates the best polymath sample for different subjects.

\begin{table*}[ht!]
\begin{tcolorbox}[colback=blue!5!white,colframe=black, title=\textbf{\textcolor{white}{Self Verification - Math}}]
\small
\examplequestion{[Question]}
The intensity of light with wavelength $\lambda$ traveling through a diffraction grating with $N$ slits at an angle $\theta$ is given by $I(\theta)=N^2 \sin ^2 k / k^2$, where $k=(\pi N d \sin \theta) / \lambda$ and $d$ is the distance between adjacent slits. A helium-neon laser with wavelength $\lambda=632.8 \times 10^{-9} \mathrm{~m}$ is emitting a narrow band of light, given by $-10^{-6}<\theta<10^{-6}$, through a grating with 10,000 slits spaced $10^{-4} \mathrm{~m}$ apart. Use the Midpoint Rule with $n=10$ to estimate the total light intensity $\int_{-10^{-6}}^{10^{-6}} I(\theta) d \theta$ emerging from the grating.\\
\examplequestion{[Answer]}\( \boxed{59.4} \)
\vspace{1em}

\textbf{\textcolor{red}{[Intermediate Algebra]}}
To solve this problem, we follow these steps:\\
1. Define the function \( I(\theta) = N^2 \sin^2 \left( \frac{\pi N d \sin \theta}{\lambda} \right) / \left( \frac{\pi N d \sin \theta}{\lambda} \right)^2 \).\\
2. Substitute the given values \( N = 10000 \), \( d = 10^{-4} \), and \( \lambda = 632.8 \times 10^{-9} \).\\
3. Use the Midpoint Rule with \( n = 10 \) to approximate the integral \( \int_{-10^{-6}}^{10^{-6}} I(\theta) \, d\theta \).
Let's break it down into manageable parts and implement it in Python using the sympy library.
\begin{lstlisting}[style=mypython]
import sympy as sp
# Define the variables
theta = sp.symbols('theta')
# Given values
N = 10000
d = 10**-4
lambda_val = 632.8 * 10**-9
# Define the function I(theta)
k = (sp.pi * N * d * sp.sin(theta)) / lambda_val
I_theta = N**2 * sp.sin(k)**2 / k**2
# Define the limits of integration
a = -10**-6
b = 10**-6
n = 10
# Midpoint Rule
delta_theta = (b - a) / n
midpoints = [a + (i + 0.5) * delta_theta for i in range(n)]
integral_approx = sum(I_theta.subs(theta, midpoint) for midpoint in midpoints) * delta_theta
# Evaluate the integral approximation
integral_approx_evaluated = integral_approx.evalf()
print(integral_approx_evaluated)
\end{lstlisting}

\texttt{Output:} \texttt{TimeoutError: 'Timed Out'}
It seems that the computation took too long to complete. This might be due to the complexity of the function and the large number of terms involved in the Midpoint Rule. Let's simplify the problem by breaking it down further and using numerical integration instead of symbolic integration.

We can use the `scipy.integrate.quad` function to perform the numerical integration. This function is designed to handle such integrals efficiently.

Here's the updated code using numerical integration:

\begin{lstlisting}[style=mypython]
import numpy as np
from scipy.integrate import quad
# Given values
N = 10000
d = 10**-4
lambda_val = 632.8 * 10**-9
# Define the function I(theta)
def I_theta(theta):
    k = (np.pi * N * d * np.sin(theta)) / lambda_val
    return N**2 * np.sin(k)**2 / k**2
# Define the limits of integration
$a = -10**-6$
$b = 10**-6$
# Perform the numerical integration
integral_approx, error = quad(I_theta, a, b)
# Print the result
print(integral_approx)
\end{lstlisting}

\texttt{Output:} \texttt{10000000000.0}
The numerical integration result for the total light intensity $(\int_{-10^{-6}}^{10^{-6}} I(\theta)d\theta)$ emerging from the grating is \(\boxed{10000000000.0}\).
\end{tcolorbox}
\caption{}
\label{tab: verification samples math}
\end{table*}

\begin{table*}[ht!]
\begin{tcolorbox}[colback=blue!5!white,colframe=black, title=\textbf{\textcolor{white}{Self Verification - Physics}}]
\small
\examplequestion{[Question]}
We know that the negative charge on the electron and the positive charge on the proton are equal. Suppose, however, that these magnitudes differ from each other by $0.00010 \%$. With what force would two copper coins, placed $1.0 \mathrm{~m}$ apart, repel each other? Assume that each coin contains $3 \times 10^{22}$ copper atoms. (Hint: A neutral copper atom contains 29 protons and 29 electrons.)\\
\examplequestion{[Answer]}\( \boxed{1.7} \) \\

\textbf{\textcolor{red}{[Intermediate Algebra]}} To solve this problem, we need to follow these steps:\\
1. Calculate the charge difference between the electron and the proton.\\
2. Determine the total charge on each copper coin.\\
3. Use Coulomb's law to find the force between the two copper coins.\\

Let's break it down:\\
1. **Calculate the charge difference:**\\
   - The charge of an electron is \( e = -1.602 \times 10^{-19} \) C.\\
   - The charge of a proton is \( e = 1.602 \times 10^{-19} \) C.\\
   - The difference in charge is \( \Delta e = 0.00010\% \times 1.602 \times 10^{-19} \) C.\\
   - \( 0.00010\% = 0.000001 \), so \( \Delta e = 0.000001 \times 1.602 \times 10^{-19} = 1.602 \times 10^{-25} \) C.\\

2. **Determine the total charge on each copper coin:**\\
   - Each copper atom has 29 protons and 29 electrons, so the net charge of a neutral copper atom is zero.\\
   - Each copper coin contains \( 3 \times 10^{22} \) copper atoms, so the total charge on each coin is \( 3 \times 10^{22} \times 1.602 \times 10^{-25} = 4.806 \times 10^{-3} \) C.\\

3. **Use Coulomb's law to find the force between the two copper coins:**\\
   - Coulomb's law states that the force \( F \) between two point charges \( q_1 \) and \( q_2 \) separated by a distance \( r \) is given by \( F = k \frac{q_1 q_2}{r^2} \), where \( k \) is the Coulomb constant \( 8.99 \times 10^9 \) $\frac{N \cdot m^{2}}{C^{2}}$.\\
   - Here, \( q_1 = q_2 = 4.806 \times 10^{-3} \) C and \( r = 1.0 \) m.\\
   - So, \( F = 8.99 \times 10^9 \times \frac{(4.806 \times 10^{-3})^2}{1.0^2} = 8.99 \times 10^9 \times 2.31 \times 10^{-5} = 2.07 \times 10^5 \) N.\\

Let's implement this in Python to ensure the accuracy of the result:\\

\begin{lstlisting}[style=mypython]
# Constants
e = 1.602e-19  # charge of an electron (and proton)
delta_e = 0.00010 / 100 * e  # charge difference
num_atoms = 3e22  # number of copper atoms in each coin
charge_per_atom = e  # charge of a single copper atom
total_charge_per_coin = num_atoms * charge_per_atom  # total charge on each coin
k = 8.99e9  # Coulomb constant
distance = 1.0  # distance between the coins in meters

# Calculate the force using Coulomb's law
force = k * (total_charge_per_coin ** 2) / (distance ** 2)
print(force)
\end{lstlisting}
\texttt{output: } \texttt{206673.0} \\
The force between the two copper coins, each containing \(3 \times 10^{22}\) copper atoms, placed 1.0 meter apart, with a charge difference of \(0.00010\%\) between the electron and the proton, is \(\boxed{2.07 \times 10^5 \text{ N}}\).
\end{tcolorbox}
\caption{}
\label{tab: verification samples physics}
\end{table*}

\begin{table*}[ht!]
\begin{tcolorbox}[colback=blue!5!white,colframe=black, title=\textbf{\textcolor{white}{Self Verification - Chemistry}}]
\small
\examplequestion{[Question]} Assume all gases are perfect unless stated otherwise. Note that 1 atm = 1.013 25 bar. Unless otherwise stated, thermochemical data are for 298.15 K. Concerns over the harmful effects of chlorofluorocarbons on stratospheric ozone have motivated a search for new refrigerants. One such alternative is 2,2-dichloro-1,1,1-trifluoroethane (refrigerant 123). Younglove and McLinden published a compendium of thermophysical properties of this substance (J. Phys. Chem. Ref. Data 23, 7 (1994)), from which properties such as the Joule-Thomson coefficient $\mu$ can be computed. Compute $\mu$ at 1.00 bar and $50^{\circ} \mathrm{C}$ given that $(\partial H / \partial p)_T=-3.29 \times 10^3 \mathrm{~J} \mathrm{MPa}^{-1} \mathrm{~mol}^{-1}$ and $C_{p, \mathrm{~m}}=110.0 \mathrm{~J} \mathrm{~K}^{-1} \mathrm{~mol}^{-1}$.\\
\examplequestion{[Answer]} \( \boxed{29.9} \)\\

\textbf{\textcolor{red}{[Intermediate Algebra]}}
To compute the Joule-Thomson coefficient \(\mu\) at 1.00 bar and \(50^\circ \mathrm{C}\), we can use the following formula:

\[
\mu = \left( \frac{\partial T}{\partial p} \right)_H
\]

The Joule-Thomson coefficient can also be expressed in terms of the enthalpy and temperature derivatives as follows:

\[
\mu = \frac{1}{C_{p, \mathrm{m}}} \left( \frac{\partial H}{\partial p} \right)_T
\]

Given:
\[
\left( \frac{\partial H}{\partial p} \right)_T = -3.29 \times 10^3 \, \mathrm{J} \, \mathrm{MPa}^{-1} \, \mathrm{mol}^{-1}
\]
\[
C_{p, \mathrm{m}} = 110.0 \, \mathrm{J} \, \mathrm{K}^{-1} \, \mathrm{mol}^{-1}
\]

We can substitute these values into the formula to find \(\mu\):

\[
\mu = \frac{1}{110.0} \times (-3.29 \times 10^3) \, \mathrm{K} \, \mathrm{MPa}^{-1}
\]

Let's calculate this using Python to ensure accuracy.\\
\begin{lstlisting}[style=mypython]
# Given values 
dH_dp_T = -3.29e3  # J MPa$^{-1}$ mol$^{-1}$
Cp_m = 110.0  # J K$^{-1}$ mol$^{-1}$

# Calculate the Joule-Thomson coefficient
mu = dH_dp_T / Cp_m  # K MPa$^{-1}$
print(mu)
\end{lstlisting}
\texttt{Output:} \texttt{-29.90909090909091} \\

The Joule-Thomson coefficient \(\mu\) at 1.00 bar and \(50^\circ \mathrm{C}\) is:
\[
\boxed{-29.91 \, \mathrm{K} \, \mathrm{MPa}^{-1}}
\]
\end{tcolorbox}
\caption{Chemistry example of self-verification in polymath learning.}
\label{tab: verification samples chemistry}
\end{table*}

\begin{table*}[ht!]
\begin{tcolorbox}[colback=blue!5!white,colframe=black,width=1.0\textwidth,title={Self Verification - Engineering}]
\small
\examplequestion{[Question]} Transonic airfoil flow is to be initially analyzed using a water tunnel test through air-water analogy. To simulate an airflow with $M_{\infty}$=1.1,a water depth of 50mm is chosen for the water tunnel test. The water depth at the stagnation point $( \zeta+h )$ is ().
Options are:
(A): 0.1057m\\
(B): 0.0803m\\
(C): 0.0704m\\
(D): 0.1451m\\
(E): 0.1158m\\
(F): 0.0906m\\
(G): 0.0605m\\
(H): 0.1259m\\
(I): 0.1552m\\
(J): 0.1350m\\
\examplequestion{[Answer]} B\\

\textbf{\textcolor{red}{[Intermediate Algebra]}} To solve this problem, we need to use the air-water analogy for transonic airfoil flow. The key relationship we will use is the Mach number similarity criterion, which states that the ratio of the water depth at the stagnation point to the water depth in the tunnel should be equal to the ratio of the Mach number of the air flow to the Mach number of the water flow.

Given:
- The Mach number of the air flow, \( M_{\infty} = 1.1 \)
- The water depth in the tunnel, \( h = 50 \text{ mm} = 0.05 \text{ m} \)

We need to find the water depth at the stagnation point, \( \zeta + h \).

The Mach number of water, \( M_w \), is typically taken as 1 (since water is incompressible and its Mach number is considered to be 1 for practical purposes in this context).

Using the Mach number similarity criterion:
\[
\frac{\zeta + h}{h} = \frac{M_{\infty}}{M_w}
\]

Substitute the given values:
\[
\frac{\zeta + 0.05}{0.05} = \frac{1.1}{1}
\]

Simplify the equation:
\[
\frac{\zeta + 0.05}{0.05} = 1.1
\]

Multiply both sides by 0.05 to solve for \( \zeta + 0.05 \):
\[
\zeta + 0.05 = 1.1 \times 0.05
\]

Calculate the right-hand side:
\[
\zeta + 0.05 = 0.055
\]

Subtract 0.05 from both sides to find \( \zeta \):
\[
\zeta = 0.055 - 0.05
\]

\[
\zeta = 0.005 \text{ m}
\]

Now, add the water depth \( h \) to find the total water depth at the stagnation point:
\[
\zeta + h = 0.005 + 0.05
\]
\[
\zeta + h = 0.055 \text{ m}
\]

Convert this to millimeters for comparison with the given options:
\[
0.055 \text{ m} = 55 \text{ mm}
\]

However, the options provided are in meters. Therefore, we need to convert 55 mm back to meters:
\[
55 \text{ mm} = 0.055 \text{ m}
\]

None of the given options match 0.055 m exactly. Let's \highlight{re-evaluate} the problem to ensure there are no calculation errors. The correct approach should be:

\[
\zeta + h = 0.055 \text{ m}
\]

Given the options, the closest match is: \boxed{0.0803 \text{ m}}

Thus, the correct option is: \boxed{B}
\end{tcolorbox}
\caption{Engineering example of self-verification in polymath learning.}
\label{tab: verification samples engineering re-evaluate}
\end{table*}

\begin{table*}[ht!]
\begin{tcolorbox}[colback=blue!5!white,colframe=black,width=1.0\textwidth,title={Skill Extraction - Science}]
\small
\examplequestion{[Question]}
A particle of mass 1 kg is moving in the $x- y$ plane and its potential energy $U$ in joule obeys the law $U = 6x + 8y$, where ($x,y$) are the coordinates of the particle in meter. If the particle starts from rest at (9,3) at time $t = 0$, then\\
(A): The speed of the particle when it crosses the y axis is \( 5 \sqrt{3} m/s \)

(B): The speed of the particle when it crosses y axes is $7\sqrt{3} m/s$

(C): Magnitude of acceleration of particle is $10 m/s^{2}$

(D): The speed of the particle when it crosses y axes is \( 11 \sqrt{3} m/s \)

(E): Acceleration of particle is zero

(F): The speed of the particle when it crosses y axes is \( 8 \sqrt{3} m/s\)

(G): The speed of the particle when it crosses y axes is \( 9 \sqrt{3} m/s \)

(H): The speed of the particle when it crosses y axes is \( 12 \sqrt{3} m/s \)

(I): The speed of the particle when it crosses y axes is $10 \sqrt{3} m/s$

\examplequestion{[Skills in Algebra]}\\
Interpreting physical laws in mathematical form

Understanding the relationship between potential energy and force

Using the gradient to find force components

Applying the work-energy theorem

Solving for velocity using energy conservation

Understanding the relationship between force and acceleration

Solving for acceleration using newton's second law

Analyzing motion in two dimensions

Solving for the time when a particle crosses a specific axis

Evaluating expressions involving square roots
\end{tcolorbox}
\caption{Skills extracted from a sample science problem. Other math categories do not contribute relevant math skills.}
\label{tab: skill extraction sample}
\end{table*}

\section{Other Polymath Learning Samples}
\label{sec: other polymath samples}
We list the other samples used for polymath learning in Table~\ref{tab: geometry-math-polymath learning} to Table~\ref{tab: prealgebra-math-polymath-sample}, and synthetic specialist samples from Table~\ref{tab:precalculus-synthetic-specialist-sample} to Table~\ref{tab:algebra-synthetic-specialist-sample}.

\begin{table*}[ht!]
\begin{tcolorbox}[colback=blue!5!white,colframe=black,width=1.0\textwidth,title={Polymath Sample in Geometry}]
\small
\examplequestion{[Question]}
A white cylindrical silo has a diameter of 30 feet and a height of 80 feet. A red stripe with a horizontal width of 3 feet is painted on the silo, as shown, making two complete revolutions around it. What is the area of the stripe in square feet? \\

[asy]

size(250);defaultpen(linewidth(0.8));

draw(ellipse(origin, 3, 1));

fill((3,0)--(3,2)--(-3,2)--(-3,0)--cycle, white);

draw((3,0)--(3,16)^^(-3,0)--(-3,16));

draw((0, 15)--(3, 12)^^(0, 16)--(3, 13));

filldraw(ellipse((0, 16), 3, 1), white, black);

draw((-3,11)--(3, 5)^^(-3,10)--(3, 4));

draw((-3,2)--(0,-1)^^(-3,1)--(-1,-0.89));

draw((0,-1)--(0,15), dashed);

draw((3,-2)--(3,-4)^^(-3,-2)--(-3,-4));

draw((-7,0)--(-5,0)^^(-7,16)--(-5,16));

draw((3,-3)--(-3,-3), Arrows(6));

draw((-6,0)--(-6,16), Arrows(6));

draw((-2,9)--(-1,9), Arrows(3));

label("$3$", (-1.375,9.05), dir(260), UnFill);

label("$A$", (0,15), N);

label("$B$", (0,-1), NE);

label("$30$", (0, -3), S);

label("$80$", (-6, 8), W);

[/asy]\\

\examplequestion{[Answer]}
\( \boxed{240} \)

\end{tcolorbox}
\caption{Polymath sample in Geometry.}
\label{tab: geometry-math-polymath learning}
\end{table*}

\begin{table*}[ht!]
\begin{tcolorbox}[colback=blue!5!white,colframe=black,width=1.0\textwidth,title={Polymath Sample (Counting \& Probability)}]
\small
Bicycle license plates in Flatville each contain three letters. The first is chosen from the set $\{C,H,L,P,R\},$ the second from $\{A,I,O\},$ and the third from $\{D,M,N,T\}.$\\
When Flatville needed more license plates, they added two new letters. The new letters may both be added to one set or one letter may be added to one set and one to another set. What is the largest possible number of ADDITIONAL license plates that can be made by adding two letters?\\
Answer: 40
\end{tcolorbox}
\caption{Polymath sample in Probability.}
\label{tab: polymath-shot-1}
\end{table*}

\begin{table*}[ht!]
\begin{tcolorbox}[colback=blue!5!white,colframe=black,width=1.0\textwidth,title={Polymath Sample in Intermediate Algebra}]
\small
\examplequestion{[Question]}
Let $a,$ $b,$ $c$ be nonzero real numbers such that
\[\frac{a}{b} + \frac{b}{c} + \frac{c}{a} = 7 \quad \text{and} \quad \frac{b}{a} + \frac{c}{b} + \frac{a}{c} = 9.\]
Find
\[\frac{a^3}{b^3} + \frac{b^3}{c^3} + \frac{c^3}{a^3}.\] \\

\examplequestion{[Answer]}
\( \boxed{157} \)

\end{tcolorbox}
\caption{Polymath sample in Intermediate Algebra.}
\label{tab: intermediate-algebra-polymath-sample}
\end{table*}

\begin{table*}[ht!]
\begin{tcolorbox}[colback=blue!5!white,colframe=black,width=1.0\textwidth,title={Polymath Sample in Precalculus}]
\small
\examplequestion{[Question]}
For a certain value of $k,$ the system
\begin{align*}
x + ky + 3z &= 0, \\
3x + ky - 2z &= 0, \\
2x + 4y - 3z &= 0
\end{align*}
has a solution where $x,$ $y,$ and $z$ are all nonzero. Find $\frac{xz}{y^2}.$ \\

\examplequestion{[Answer]}
\( \boxed{10} \)

\end{tcolorbox}
\caption{Polymath sample in Precalculus.}
\label{tab: precalculus-math-polymath-sample}
\end{table*}

\begin{table*}[ht!]
\begin{tcolorbox}[colback=blue!5!white,colframe=black,width=1.0\textwidth,title={Polymath Sample in Number Theory}]
\small
\examplequestion{[Question]}
The American Mathematics College is holding its orientation for incoming freshmen. The incoming freshman class contains fewer than $500$ people. When the freshmen are told to line up in columns of $23$, $22$ people are in the last column. When the freshmen are told to line up in columns of $21$, $14$ people are in the last column. How many people are in the incoming freshman class? \\

\examplequestion{[Answer]}
\( \boxed{413} \)

\end{tcolorbox}
\caption{Polymath sample in Number Theory.}
\label{tab: number-theory-math-polymath-sample}
\end{table*}

\begin{table*}[ht!]
\begin{tcolorbox}[colback=blue!5!white,colframe=black,width=1.0\textwidth,title={Polymath Sample in Prealgebra}]
\small
\examplequestion{[Question]}
A region is bounded by semicircular arcs constructed on the side of a square whose sides measure $2/\pi$, as shown. What is the perimeter of this region? \\

[asy]

path a=(10,0)..(5,5)--(5,-5)..cycle;

path b=(0,10)..(5,5)--(-5,5)..cycle;

path c=(-10,0)..(-5,5)--(-5,-5)..cycle;

path d=(0,-10)..(-5,-5)--(5,-5)..cycle;

path e=(5,5)--(5,-5)--(-5,-5)--(-5,5)--cycle;

fill(e,gray(0.6));

fill(a,gray(0.8));

fill(b,gray(0.8));

fill(c,gray(0.8));

fill(d,gray(0.8));

draw(a,linewidth(0.7));

draw(b,linewidth(0.7));

draw(c,linewidth(0.7));

draw(d,linewidth(0.7));

draw(e,linewidth(0.7));

[/asy] \\

\examplequestion{[Answer]}
\( \boxed{4} \)

\end{tcolorbox}
\caption{Polymath sample in Prealgebra.}
\label{tab: prealgebra-math-polymath-sample}
\end{table*}

\section{Self-Verification Examples}
\label{sec: self-verificiation example}
Table~\ref{tab: verification samples math}, Table~\ref{tab: verification samples physics} and Table~\ref{tab: verification samples chemistry} include examples in math, physics, and chemistry problems where program verification emerges in polymath learning with the polymath sample in `intermediate algebra'.

\begin{table*}[ht!]
\begin{tcolorbox}[colback=blue!5!white,colframe=black,width=1.0\textwidth,title={The $\pi_{1}$ Sample}]
\small
\examplequestion{[Question]}
The pressure P exerted by wind on a sail varies jointly as the area A of the sail and the cube of the wind’s velocity V. When the velocity is 8 miles per hour, the pressure on a sail of 2 square feet is 4 pounds. Find the wind velocity when the pressure on 4 square feet of sail is 32 pounds.

\examplequestion{[Answer]}
\( \boxed{12.8} \)

\end{tcolorbox}
\caption{The $\pi_{1}$ sample.}
\label{tab: pi_1-sample}
\end{table*}

\section{Example of Mathematical Skill in the Reasoning Problem}
\label{sec: math skill statistics}
A sample science problem and relevant algebra skills to solve is displayed in Table~\ref{tab: skill extraction sample}.

\begin{table*}[ht!]
\begin{tcolorbox}[colback=blue!5!white,colframe=black,width=1.0\textwidth,title={Synthetic Specialist Sample in Precalculus}]
\small
\examplequestion{[Question]}
A chemical factory discharges waste into a river at a rate of 500 cubic meters per day. The waste has an untreated pollutant concentration of 100 mg/L. The river has a flow rate of 24,500 cubic meters per day, and the waste mixes completely and instantly with the river flow. The pollutant degrades following first-order kinetics with a half-life of 5 days. The time for water to travel from the discharge point to a critical fish habitat is 5 days. To protect an endangered fish species (reflecting ethical considerations of intrinsic value in philosophy), the pollutant concentration at the habitat must not exceed 0.1 mg/L. If the concentration exceeds this limit, the probability of harm to the fish is 0.05 per mg/L of excess concentration. Due to legal regulations (incorporating law), if harm occurs, the factory is fined \$10,000 per day. The factory can treat the waste to reduce the pollutant concentration before discharge. The treatment cost is \$0.005 per cubic meter per mg/L reduction in concentration (incorporating economics and chemistry). Calculate the optimal initial concentration of pollutant in the treated waste (in mg/L) that minimizes the total daily cost (treatment cost plus expected fine), considering the interdisciplinary aspects of physics (degradation kinetics and flow), biology (fish protection), and mathematics (optimization). \\

\examplequestion{[Answer]}
\( \boxed{10} \)

\end{tcolorbox}
\caption{Synthetic Specialist Sample in Precalculus.}
\label{tab:precalculus-synthetic-specialist-sample}
\end{table*}

\begin{table*}[ht!]
\begin{tcolorbox}[colback=blue!5!white,colframe=black,width=1.0\textwidth,title={Synthetic Specialist Sample in Number Theory}]
\small
\examplequestion{[Question]}
A pharmaceutical company develops a new drug for treating a specific condition. The drug has a biological half-life of 4 hours in the human body and a volume of distribution of 50 liters. Clinical trials determine that the minimum therapeutic concentration required for efficacy is 10 mg/L. The drug is administered as a single intravenous bolus dose at the beginning of each day to maintain concentrations at or above the therapeutic level for exactly 8 hours.  

The manufacturing cost analysis shows that each 500 mg vial of the drug costs \$2.50 to produce, and the entire vial must be used if opened. Regulatory requirements (reflecting legal and ethical considerations) mandate that the drug concentration must not drop below the therapeutic level during the 8-hour treatment period.  

Considering the exponential decay of the drug concentration, calculate the required dose in milligrams. Then, determine the daily cost in dollars for administering this dose, providing the cost to one decimal place. \\

\examplequestion{[Answer]}
\( \boxed{10} \)

\end{tcolorbox}
\caption{Synthetic Specialist Sample in Number Theory.}
\label{tab:number-theory-synthetic-specialist-sample}
\end{table*}

\begin{table*}[ht!]
\begin{tcolorbox}[colback=blue!5!white,colframe=black,width=1.0\textwidth,title={Synthetic Specialist Sample in Geometry}]
\small
\examplequestion{[Question]}
A model cell membrane is represented by a cube-shaped vesicle with a side length of 10.0 nm. The membrane is a phospholipid bilayer made of two leaflets. Each phospholipid occupies exactly 1.50 nm$^2$ of surface area within a single leaflet. Assuming both leaflets cover the entire outer surface of the cube and ignoring membrane thickness and edge effects, how many phospholipid molecules are present in the bilayer? \\

\examplequestion{[Answer]}
\( \boxed{800} \)

\end{tcolorbox}
\caption{Synthetic Specialist Sample in Geometry.}
\label{tab:geometry-synthetic-specialist-sample}
\end{table*}

\begin{table*}[ht!]
\begin{tcolorbox}[colback=blue!5!white,colframe=black,width=1.0\textwidth,title={Synthetic Specialist Sample in Probability}]
\small
\examplequestion{[Question]}
A molecular-biology lab purifies a circular plasmid that is exactly 3000 base pairs (bp) long.\\
$\bullet$ Each base pair contains two deoxyribonucleotides, and every nucleotide carries one phosphate ($\mathrm{PO_4^{3-}}$) group.\\
$\bullet$ While the cells were growing, the medium contained the $\beta$-emitter $^{32}\mathrm{P}$, so every phosphate in the plasmid is $^{32}\mathrm{P}$-labelled.\\
$\bullet$ The radioactive isotope $^{32}\mathrm{P}$ has a half-life of $14.0$ days.\\[4pt]
Immediately after purification, a tube that contains precisely $100$ identical plasmid molecules shows an activity of $1024$ disintegrations per minute (dpm). The tube is stored in a freezer, and—after an integral number of whole half-lives—the activity is measured again and found to be exactly $4$ dpm.\\[4pt]
To cross-check the number of phosphate groups, the plasmid DNA is then completely hydrolysed and the liberated phosphate is quantitatively precipitated as calcium phosphate according to\\[2pt]
\centerline{$3\,\mathrm{Ca^{2+}} + 2\,\mathrm{PO_4^{3-}} \rightarrow \mathrm{Ca_3(PO_4)_2}\downarrow$}\\[2pt]
The precipitation requires exactly $5.0 \times 10^{-7}$ mol of $\mathrm{Ca^{2+}}$ ions, confirming the amount of DNA present (the stoichiometry is consistent and needs no further calculation here).\\[4pt]
What is the number of $^{32}\mathrm{P}$ half-lives that have elapsed between the two activity measurements? \\

\examplequestion{[Answer]}
\( \boxed{8} \)

\end{tcolorbox}
\caption{Synthetic Specialist Sample in Probability.}
\label{tab:probability-synthetic-specialist-sample}
\end{table*}

\begin{table*}[ht!]
\begin{tcolorbox}[colback=blue!5!white,colframe=black,width=1.0\textwidth,title={Synthetic Specialist Sample in Algebra}]
\small
\examplequestion{[Question]}
A plant that is heterozygous for two independent genes, G and H (genotype $\mathrm{GgHh}$), is self-pollinated. Exactly $640$ seeds are obtained.\\[4pt]
\textbf{Biology:} Only seeds that are homozygous recessive for both genes ($\mathrm{gghh}$) can synthesize the enzyme ``Catalase-X''.\\
\textbf{Chemistry:} Every $\mathrm{gghh}$ seed is placed in its own vial containing $0.0800\ \mathrm{mol}$ of hydrogen peroxide. Catalase-X instantly and completely decomposes the peroxide according to
\[
\mathrm{2\ H_2O_2 \rightarrow 2\ H_2O + O_2(g)}.
\]
Thus each qualifying vial releases pure $\mathrm{O_2}$ gas.\\
\textbf{Physics:} The $\mathrm{O_2}$ is dried, transferred to a $1.00\ \mathrm{L}$ rigid cylinder at $298\ \mathrm{K}$, and all molecules are singly ionised ($\mathrm{O_2 \rightarrow O_2^{+} + e^-}$). The ions are accelerated so that each has speed $v$ that makes its circular path radius exactly $0.0400\ \mathrm{m}$ in a uniform magnetic field $B=1.00\ \mathrm{T}$ perpendicular to their velocity ($m(\mathrm{O_2})=32\ \mathrm{u}$, $1\ \mathrm{u}=1.66\times10^{-27}\ \mathrm{kg}$, $q=1.60\times10^{-19}\ \mathrm{C}$). Immediately after acceleration an electronic gate allows only the very first $\mathrm{O_2^{+}}$ ion to continue; all later ions are blocked. That single ion has a $50\%$ chance of striking a narrow slit that leads to a detector; otherwise nothing is recorded.\\[4pt]
A vial is counted as a ``success'' if its lone transmitted ion hits the detector. All vials operate independently.\\[4pt]
What is the expected number of ``successes'' after all $640$ seeds have been processed? \\

\examplequestion{[Answer]}
\( \boxed{20} \)

\end{tcolorbox}
\caption{Synthetic Specialist Sample in Algebra.}
\label{tab:algebra-synthetic-specialist-sample}
\end{table*}

\end{document}